\documentclass{article}

\PassOptionsToPackage{numbers, compress}{natbib}

\usepackage[preprint]{neurips_2025}
\pdfoutput=1  

\usepackage[utf8]{inputenc} 
\usepackage[T1]{fontenc}    	
\usepackage{url}            		
\usepackage{booktabs}       	
\usepackage{amsfonts}       	
\usepackage{nicefrac}       	
\usepackage{microtype}      	
\usepackage{xcolor}         		
\usepackage{amsmath}
\usepackage{mathrsfs}
\usepackage{amssymb}
\usepackage{subcaption}
\usepackage{graphicx}
\usepackage{algorithm}
\usepackage{algorithmic}
\usepackage{makecell} 
\usepackage{natbib}
\usepackage{lipsum}
\usepackage{graphicx}
\usepackage{listings}
\usepackage{tcolorbox}
\usepackage{multirow}
\usepackage{tikz}
\usepackage{wrapfig}
\usepackage{amsmath}
\usetikzlibrary{arrows.meta, shapes.geometric, calc, positioning}
\usepackage[colorlinks,linkcolor=magenta,citecolor=blue, pagebackref=true,backref=false]{hyperref}

\newtheorem{theorem}{Theorem}[section]

\newtheorem{lemma}[theorem]{Lemma}
\newtheorem{proposition}[theorem]{Proposition}

\newtheorem{remark}[theorem]{Remark}
\newtheorem{assumption}[theorem]{Assumption}

\newcommand{\textproc}[1]{\textsc{#1}}

\newcommand{\ba}{\begin{array}}
\newcommand{\ea}{\end{array}}

\title{Displacement-Sparse Neural Optimal Transport}

\author{%
Peter L. Chen$^{\spadesuit,\diamondsuit}$ \quad Yue Xie$^\diamondsuit$ \quad Qingpeng Zhang$^{\diamondsuit,\clubsuit}$ \\
Department of Mathematics, Columbia University$^\spadesuit$ \\
HKU Musketeers Foundation Institute of Data Science$^\diamondsuit$ \\
HKU-Shanghai Intelligent Computing Research Center$^\clubsuit$ \\ 
\texttt{lc3826@columbia.edu}\quad \texttt{\{yxie,qpzhang\}@hku.hk}\\
}

\begin{document}

\maketitle

\begin{abstract}
Optimal transport (OT) aims to find a map $T$ that transports mass from one probability measure to another while minimizing a cost function. Recently, neural OT solvers have gained popularity in high-dimensional biological applications such as drug perturbation, due to their superior computational and memory efficiency compared to traditional exact Sinkhorn solvers. However, the overly complex high-dimensional maps learned by neural OT solvers often suffer from poor interpretability. Prior work \cite{Cuturi23} addressed this issue in the context of exact OT solvers by introducing \emph{displacement-sparse maps} via designed elastic cost, but such method failed to be applied to neural OT settings. In this work, we propose an intuitive and theoretically grounded approach to learning \emph{displacement-sparse maps} within neural OT solvers. Building on our new formulation, we introduce a novel smoothed $\ell_0$ regularizer that outperforms the $\ell_1$-based alternative from prior work. Leveraging Input Convex Neural Network's flexibility, we further develop a heuristic framework for adaptively controlling sparsity intensity—an approach uniquely enabled by the neural OT paradigm. We demonstrate the necessity of this adaptive framework in large-scale, high-dimensional training, showing not only improved accuracy but also practical ease of use for downstream applications.

\end{abstract}

\section{Introduction}

Optimal transport (OT) aims to find a map \( T \) that transfers mass from one measure to another according to a ground-truth cost. This technique has been widely applied in various machine-learning-based single-cell biology studies~\cite{Bunne23, chen2025geometricframework3dcell,chen2024w1solver, Dem22, Huizing22, klein2024moscot, Liu2023Cellstitch, Sch19, tronstad2025multistageot, wang2023cmot, Zhang21,zhu2025unbalancedot}. Compared to the traditional exact OT solver based on the Sinkhorn algorithm~\cite{Cut13}, recent studies in single-cell biology favor neural OT solvers due to their efficient scalability for large-scale and high-dimensional biological datasets, offering advantages in both computational speed and memory usage. Among these neural approaches, CellOT~\cite{Bunne23} is a prominent example that leverages the Input Convex Neural Network (ICNN) to successfully model drug perturbation problems.
\begin{figure}[t]
    \centering
    \includegraphics[width=0.95\textwidth]{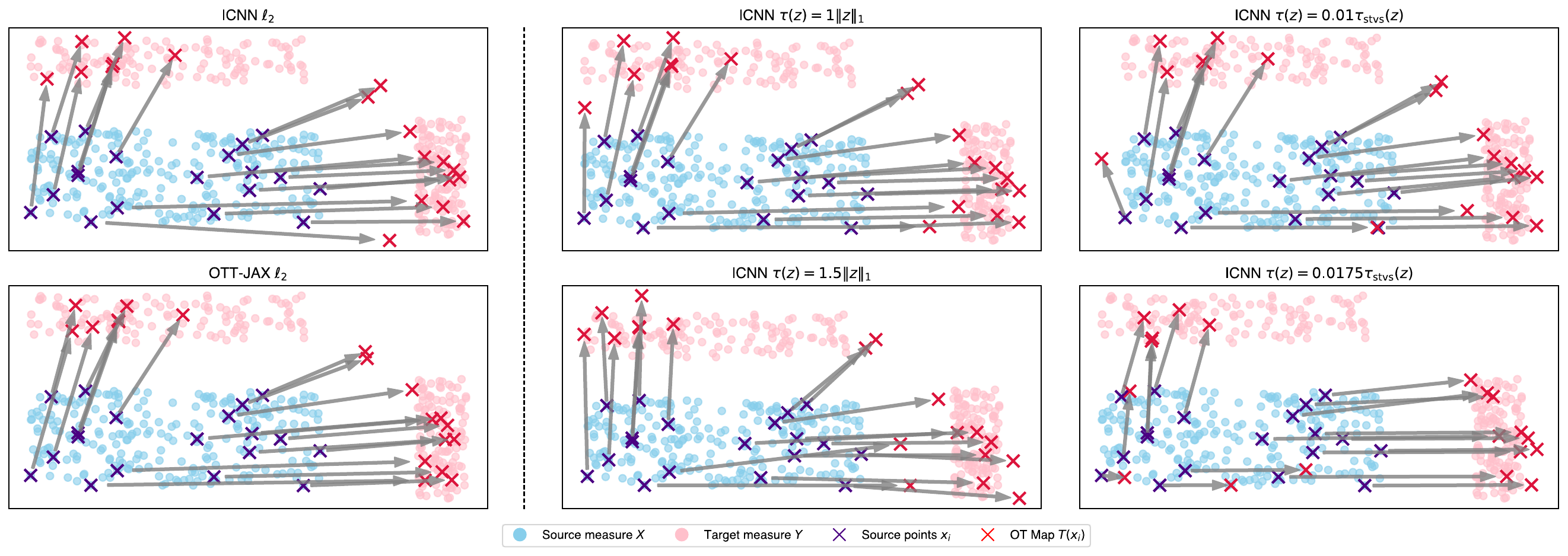} 
    \caption{OT maps induced by various sparsity penalties $\tau(\cdot)$ at different intensity levels $\lambda$. The first column presents the maps produced by ICNN OT and the traditional Sinkhorn solver implemented using \texttt{OTT-JAX} \cite{cuturi2022ottjax}, both without sparsity penalties. The subsequent columns illustrate maps learned via ICNN with sparsity penalties, specifically $\ell_{1}$ and $\tau_{\text{stvs}}$: the top row corresponds to maps trained with lower sparsity intensities, while the bottom row represents maps trained with higher sparsity intensities. Randomly selected points from the source distribution are used to showcase the displacement vectors learned through OT. Example of aggregate results are shown in \hyperref[f2]{Figure 2}.}
    \label{f1}
\end{figure}

While neural OT solvers can produce feasible maps, the lack of sparsity in the displacement vector reduces their interpretability, resulting in overly complex OT mappings that involve all dimensions of genes or cell features. Similar limitation is also present in exact Sinkhorn solvers, and to solve this problem, \citet{Cuturi23} introduced the concept of \emph{displacement-sparse maps} specifically for exact OT solvers. These maps offer improved interpretability with lower-dimensional displacement vectors while preserving their applicability in the original feature space without relying on dimensionality reduction techniques. Still, to date, no study has achieved \emph{displacement-sparse maps} learned from neural OT solvers—an area that is receiving growing attention due to its promising applicability.

\begin{wrapfigure}{r}{0.39\textwidth}
  \centering
  \includegraphics[width=0.38\textwidth]{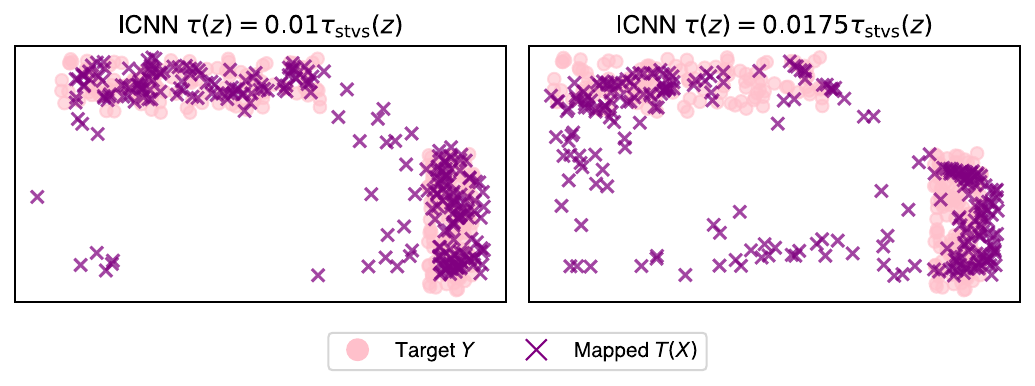}
  \caption{Aggregate results for $\tau_{\text{stvs}}$ at different sparsity intensity levels. A comparison between the mapped source elements $T(X)$ and the target elements $Y$ is shown.}
  \label{f2}
\end{wrapfigure}

Furthermore, inducing displacement sparsity in OT maps involves an inherent trade-off: as the sparsity level increases, the accuracy of the map's feasibility decreases, meaning that the source measure cannot be effectively transported to the target measure. This effect is evident in \hyperref[f2]{Figure~2}, where at higher sparsity intensity level, fewer points can be transported to the target measure, and the displacement vectors become shorter due to the stronger sparsity penalty. Finding a dynamic balance for this trade-off is practically infeasible for exact Sinkhorn OT solvers, which usually require a deterministic objective function. However, it presents a promising opportunity for neural OT solvers, which approximate the transport map iteratively over mini-batches of measures and can thus be adapted to support a adaptive sparsity-control framework.

In this work, we propose a formulation based on ICNN to learn \emph{displacement-sparse maps} for neural OT solvers and introduce a adaptive heuristic framework to control the sparsity intensity during the ICNN training. For low-dimensional tasks, where the target dimensionality is often unclear, our framework allows users to navigate the trade-off between mapping's feasibility and the degree of sparsity. In contrast, for high-dimensional tasks with a well-defined target dimensionality, we directly constrain the dimensionality of the displacement vectors, aiming to find the most feasible transport map that satisfies the dimensionality constraint. Our \textbf{contributions} are summarized below:

\textbf{(i)} We propose a novel and intuitive regularized minimax formulation for the OT dual potentials in general-purpose elastic costs, without being restricted by the convexity or differentiability of the regularizer. We provide the theoretical analysis to justify the rigorousness of incorporating dual potential regularization within the multi-layer minimax OT training.

\textbf{(ii)} We introduce a novel smoothed $\ell_{0}$ norm \cite{10.1007/978-3-540-74494-8_49} as the sparsity regularizer, which is not applicable to \citet{Cuturi23} due to its non-convexity. Through experiments, we demonstrate its superiority over the sparsity penalty norm proposed by \citet{Cuturi23} in neural OT setting.

\textbf{(iii)} We implement a heuristic framework to adaptively control sparsity intensity during training, offering two tailored strategies for end-users in both low and high dimensional tasks. Additionally, we show the necessity of using a adaptive objective function in large-scale, high-dimensional neural OT training to better align the learned displacement dimensionality with the target dimensionality.

\section{Related Works}\label{s2}

\textbf{Neural Optimal Transport.} Neural OT solvers employ continuous methods to estimate transportation plans or dual potentials through neural networks. Popular methods~\cite{alvarez-melis2022optimizing, Bunne22, pmlr-v139-fan21d, korotin2021wasserstein, Makkuva20, mokrov2021largescale, taghvaei2019} reparameterize the transportation plan or dual potentials and leverage ICNN \cite{Amos17} for approximation. In addition, \cite{Korotin21} provides a benchmark for evaluating the performance of current neural OT solvers. In line with CellOT~\cite{Bunne23}, we utilize the minimax formulation of dual potentials from \cite{Makkuva20} to introduce displacement sparsity. Comparing to other neural OT solvers, this formulation was chosen because it provides a straightforward approach to recovering displacement vectors through dual potentials. 

\textbf{Neural OT: Advantage \& Disadvantage.} Unlike exact Sinkhorn OT solvers, which rely on entropic regularization and fixed cost functions, neural OT solvers can dynamically learn transport maps for varying measures and generalize across different loss functions. Their continuous optimization through batch training enables significantly better scalability to large-scale, high-dimensional datasets in terms of both computational and memory efficiency. However, as approximation methods, neural OT solvers typically exhibit \textbf{reduced mapping accuracy} compared to exact solvers, sacrificing precision in solving ground-truth map in exchange for improved speed and memory efficiency.

\textbf{Displacement-Sparse Optimal Transport.} \citet{Cuturi23} laid the groundwork for incorporating sparsity into displacement vectors in OT by introducing a cost based on proximal operators of sparsity penalties \(\tau\) in the elastic cost to recover the OT map \(T\). Building on this, \citet{klein2024learning} extended this framework to learn transport maps under general elastic costs in exact Sinkhorn OT solvers. In the context of neural OT, our work generalizes this formulation to support arbitrary sparsity penalties, even in cases where proximal operators are undefined or intractable.

\textbf{Neural OT with General Cost.} Neural OT formulations are typically limited to quadratic cost functions, due to the difficulty of reparameterizing dual potentials for general costs. Several works~\cite{asadulaev2024neural, buzun2024enot, fan2021scalable, uscidda2023monge} have explored neural OT with more general cost functions. However, their frameworks impose restrictive assumptions such as convexity and differentiability of the cost. A more detailed discussion of these works is provided in \hyperref[a1]{Appendix~A}. These restrictions render the sparsity penalities proposed in \citet{Cuturi23} inapplicable within their settings. In contrast, our formulation supports truly general elastic cost—without any assumptions and restrictions on convexity and smoothness.

\section{Preliminaries}
\subsection{Optimal Transport}\label{sec3.1}
Given two probability measure \( P \) and \( Q \) in \( \mathbb{R}^d \) and quadratic transportation cost, the \citet{Monge1781} problem seeks to find the transport map \( T: \mathbb{R}^d \rightarrow \mathbb{R}^d \) that minimizes the transportation cost:
\[
T^* = \underset{T: T_{\#}P = Q}{\arg\min}\, \frac{1}{2} \mathbb{E}_{X \sim P} \| T(X) - X \|^2. \tag{1} \label{e1}
\]
Here, \( T^* \) is the optimal transport map among all maps \( T \) that push forward \(P\) to \(Q\). However, solving (\hyperref[e1]{1}) is challenging because the set of feasible maps \( T \) may not be convex or might not exist at all. To address this, Kantorovich relaxed the problem by allowing mass splitting, replacing the direct map \(T\) with a transportation plan coupling \(\pi\) between \(P\) and \(Q\):
\[
W_2^2(P, Q) = \inf_{\pi \in \Pi(P, Q)} \frac{1}{2} \mathbb{E}_{(X, Y) \sim \pi} \|X - Y\|^2,
\tag{2} \label{e2}
\]
where \(\Pi(P, Q)\) is the set of all couplings between \(P\) and \(Q\). This relaxation makes the problem convex and solvable via linear programming, with the optimal value being the squared 2-Wasserstein distance. To recover the map \( T \) from a coupling \(\Pi\), one can use Kantorovich's dual formulation \cite{villani2003topics}:
\[
W_2^2(P, Q) = \sup_{(f, g) \in \Phi_c} \left( \mathbb{E}_P[f(X)] + \mathbb{E}_Q[g(Y)] \right), \tag{3} \label{e3}
\]
where $\Phi_c =\left\{ (f, g) \in L^1(P) \times L^1(Q) : f(x) + g(y) \leq c(x,y) \right\}$ is the constraint space for the dual potentials \(f\) and \(g\), with \(L^1(P)\) denoting the set of integrable functions with respect to \(P\), defined as $\{f : f \text{ is measurable and } \int |f| \, dP < \infty \}\), and \(c(x,y)\) represents the transportation cost. 

Given the quadratic cost \( c(x, y) = \frac{1}{2} \| x - y \|_{2}^2 \), under \hyperref[a3.1]{Assumption 3.1} and the saturated condition \( f(x) + g(y) = \frac{1}{2} \| x - y \|_{2}^2 \), map \( T \) from $Q$ to $P$ can be recovered via $y - \nabla f^*(y)$ \cite{brenier1991polar}, where $f^*(y) = \sup_{x \in \mathbb{R}^n} (\langle x, y \rangle - f(x))$ denotes the convex conjugate of $f(y)$. The dual potentials in the constraint of (\hyperref[e3]{3}) can be further reparameterized as functions in convex space. Note that this reparameterization can only be performed under the quadratic cost function, implying that \citet{Cuturi23}'s sparsity-regularized elastic cost cannot be applied here. Eventually, (\hyperref[e3]{3}) can be reformulated into the following minimax formulation \citep[Theorem 3.3]{Makkuva20}:
\[
W_2^2(P, Q) = \sup_{\substack{f \in \mathtt{CVX}(P) \\ f^* \in L^1(Q)}} \inf_{g \in \mathtt{CVX}(Q)} \mathcal{V}_{P,Q}(f, g) + C_{P,Q}, \tag{4} \label{e4}
\]
where \(\mathcal{V}_{P,Q}(f, g) = -\mathbb{E}_P[f(X)] - \mathbb{E}_Q[\langle Y, \nabla g(Y)\rangle \) \(- f(\nabla g(Y))]\),
$C_{P,Q} = \frac{1}{2} \mathbb{E}[\|X\|^2 + \|Y\|^2]$ is a constant, and $\mathtt{CVX}(P)$ denotes the set of all convex functions in $L^1(P)$. The new minimax formulation parameterizes $f$ and $g$ into the convex space using a separate alternating optimization scheme, which can be learned via ICNN. 

\subsection{Input Convex Neural Network} \label{s3.2}
\begin{figure}[ht]
    \centering
    \begin{tikzpicture}[
        >={Stealth},
        thick,
        node distance=0.8cm and 0.8cm,
        every node/.style={draw, rectangle, minimum height=0.4cm, minimum width=0.5cm, font=\scriptsize},
        every path/.style={->}
      ]

      \node (y) {$y$};
      \node[right=of y] (z1) {$z_1$};
      \node[right=of z1] (z2) {$z_2$};
      \node[right=of z2] (z3) {$z_3$};
      \node[right=of z3] (dots) {$\cdots$};
      \node[right=of dots] (zk) {$z_k$};

      \draw[->] (y) -- (z1) node[midway, above, draw=none] {$W_0^{(y)}$};
      \draw[->] (z1) -- (z2) node[midway, above, draw=none] {$W_1^{(z)}$};
      \draw[->] (z2) -- (z3) node[midway, above, draw=none] {$W_2^{(z)}$};
      \draw[->] (z3) -- (dots);
      \draw[->] (dots) -- (zk);

      \draw[->] (y) |- ([yshift=-0.4cm] z2.south) -- (z2.south) node[pos=0.55, right, draw=none] {$W_1^{(y)}$};
      \draw[->] (y) |- ([yshift=-0.4cm] z3.south) -- (z3.south) node[pos=0.55, right, draw=none] {$W_2^{(y)}$};
      \draw[->] (y) |- ([yshift=-0.4cm] zk.south) -- (zk.south) node[pos=0.55, right, draw=none] {$W_{k-1}^{(y)}$};

    \end{tikzpicture}
    \caption{Input Convex Neural Network (ICNN) Structure.}
    \label{f3}
\end{figure}

\citet{Amos17} introduced Input Convex Neural Networks (ICNNs) to model convex functions by optimizing a set of parameters, \(\{W_{i}^{(y)}, W_{i}^{(z)}\}\), during each iteration. Given an initial input \(y\), the ICNN learns the convex function \(z_k\) through the following formulation:
\[
z_{i+1} = g_i \left( W_i^{(z)} z_i + W_i^{(y)} y + b_i \right), \quad f(y; \theta) = z_k,
\]
where \(\theta = \left\{ W_{0:k-1}^{(y)}, W_{1:k-1}^{(z)}, b_{0:k-1} \right\}\) denotes the model parameters and \(g_i\) is a non-linear activation function. During each iteration, a small batch is drawn, and the ICNN learns \(f\) and \(g\) within \(\mathtt{ICNN}(\mathbb{R}^d)\) by optimizing the minimax problem with parameters \(\theta_f\) and \(\theta_g\).

To ensure the convexity of the network, weights $W_{i}^{(y)}$ and $W_{i}^{(z)}$ must remain non-negative. \citet{Makkuva20} demonstrated the feasibility of adding an arbitrary bounds to \(\theta\) during the training process; This implementation indicates that \(\theta_f\) and \(\theta_g\) can be learned from a compact and uniformly bounded function space $\mathtt{ICNN}(\mathbb{R}^d)$ given $P,Q\in \mathbb{R}^d$, which could simplify the original convex function space by restricting it to a smaller subspace compatible with $\mathtt{ICNN}(\mathbb{R}^d)$:
\begin{assumption}\label{a3.1}
We assume that both source measure $P$ and target measure \(Q\) admits a density with respect to the Lebesgue measure, and that there exist convex subsets \(\mathcal{S}(P) \subset \mathtt{CVX}(P)\) and \(\mathcal{S}(Q) \subset \mathtt{CVX}(Q)\) which are uniformly bounded. During the ICNN training, we have \(
\mathtt{ICNN}(\mathbb{R}^d) \subset \mathcal{S}(P) \) and \(
\mathtt{ICNN}(\mathbb{R}^d) \subset \mathcal{S}(Q).\)
\end{assumption}

\begin{proposition}
\label{prop:existence-opt-map}
Under \hyperref[a3.1]{Assumption 3.1}, there exists an optimal solution \((f_0, g_0)\). Furthermore, the corresponding optimal transport map \(T\) from \(Q\) to \(P\) can be directly recovered via \(\nabla g_0\).
\end{proposition}

\subsection{Sparsity-inducing Penalties $\tau$}
\label{sec3.3}
\citet{Cuturi23} proposed the following cost function to induce displacement sparsity, with $\lambda$ as sparsity intensity and $\tau$ to be specific sparsity penalty:
\[
c(\mathbf{x}, \mathbf{y}) = \frac{1}{2} \| \mathbf{x} - \mathbf{y} \|_{2}^2 + \lambda \cdot \tau(\mathbf{y} - \mathbf{x}). \tag{5} \label{e5}
\]
Under this regularized cost, the optimal transport mapping can be recovered via:
\[
T(\mathbf{x}) = \mathbf{x} - \mathrm{prox}_{\tau} \left( \mathbf{x} - \sum_{j=1}^m p^j(\mathbf{x}) \left( \mathbf{y}^j + \nabla\tau(\mathbf{x} - \mathbf{y}^j) \right) \right),
\]
where $\mathrm{prox}_{\tau}$ denotes the proximal operator of $\tau$, and $p^j(\mathbf{x})$ represents the $\mathbf{x}$-dependent Gibbs distribution over the $m$-simplex. \citet{Cuturi23} applied the following penalty functions $\tau$,  the $\ell_{1}$-norm and Vanishing Shrinkage STVS ($\tau_{\text{stvs}}$):

\textbf{\(\ell_{1}\)-Norm.} The \(\ell_{1}\)-norm directly penalizes the magnitude of the displacement vector, encouraging it to move towards the nearest target and inducing sparsity.

\textbf{Vanishing Shrinkage STVS.} \citet{schreck2015shrinkage} introduce the soft-thresholding operator with vanishing shrinkage for displacement vector $\mathbf{z}$, with \(\sigma(\mathbf{z})=\operatorname{asinh}(\frac{z}{2\gamma})\) as element-wise operations:
\[
\tau_{\text{stvs}}(\mathbf{z}) = \gamma^2 \mathbf{1}_d^T \left( \sigma(\mathbf{z}) + \frac{1}{2} - \frac{1}{2} e^{-2 \sigma(\mathbf{z})} \right).
\]

However, both the $\ell_1$-norm and $\tau_{\text{stvs}}$ fail to directly reflect the dimensionality of the displacement vector. To address this, we introduce the most straightforward dimensionality penalty, the $\ell_{0}$-norm, into our model. Empirically, we find that its smoothed version performs better in the ICNN training.

\textbf{Smoothed $\ell_{0}$-Norm.} To approximate the discontinuous $\ell_{0}$-norm, \citet{10.1007/978-3-540-74494-8_49} proposed replacing the indicator function $\nu(\mathbf{z}) = 1_{\{\mathbf{z} \neq 0\}}$ with a smooth Gaussian function $f_{\xi}(\mathbf{z})$:
\[
\|\mathbf{z}\|_{0,\xi} = n - \sum_{i=1}^n f_{\xi}(\mathbf{z}_i) = \sum_{i=1}^n \left(1 - \exp\left(-\frac{\mathbf{z}_i^2}{2\xi^2}\right)\right).
\]
Note that the smoothed $\ell_{0}$-norm is non-convex, indicating its proximal operator may not be well-defined and therefore cannot be directly applied to (\hyperref[e5]{5}) to recover the mapping \(T\).

\section{Main Results}\label{s4}
\textbf{Motivation \& Formulation.} As explained in \S\hyperref[sec3.1]{3.1}, the dual–potential re‑parameterization in~(\hyperref[e4]{4}) fails to hold once the quadratic cost is replaced by the elastic cost~(\hyperref[e5]{5}).  This quadratic cost assumption is a well‑known bottleneck of ICNN‑based neural OT solvers, preventing them from accommodating richer cost families.  Although earlier work (\S\hyperref[s2]{2}) extends the framework to more general costs through \emph{unbiased} reformulations, it still imposes restrictive conditions on the initial cost parameterization.

Motivated by these limitations, we intentionally adopt a \emph{biased} formulation and propose the following regularized Wasserstein objective:
\begin{align}
\widetilde{W_2^2}(P,Q)
  &= C_{P,Q}
     + \sup_{f \in \mathcal{S}(P)}
       \inf_{g \in \mathcal{S}(Q)}\big(
         \mathcal{V}_{P,Q}(f,g)
     + \lambda \int_{\mathbb{R}^d}
         \tau\!\bigl(\nabla g(y)-y\bigr)\,dQ\big),
  \tag{\hyperref[e6]{6}}\label{e6}
\end{align}
where \(\nabla g(y) - y\) represents displacement vector derived from (\hyperref[e4]{4}), and \(\lambda\) is a parameter that controls the intensity of regularization. This admits a far broader class of sparsity penalties~$\tau$, including those that lack a proximal operator, thereby extending well beyond the elastic‑cost setting of~(\hyperref[e5]{5}). 

\textbf{Minimax Regularization.} Introducing regularization into a \emph{neural-parameterized minimax} problem is fundamentally different from adding it to a single–layer network.  
Here, one simultaneously optimizes over \emph{two} function spaces, so any penalty could potentially reshape the objective landscape and the intricate interaction between the dual potentials $f$ and $g$. Consequently, the potentials $f$ and $g$ \emph{may no longer correspond} to the OT dual solutions. Therefore, regularizing the dual potentials in a minimax framework must be rigorously justified both theoretically and empirically, as it modifies the equilibrium objective, alters gradient dynamics, and changes the nature of the solution itself.

\textbf{Theoretical Guarantees.} We aim to demonstrate the validity of our formulation through the following: \textbf{(i)} the regularization does not compromise the theoretical integrity of the dual formulation (\hyperref[t4.3]{Theorem 4.3}, \hyperref[t4.4]{Theorem 4.4}); \textbf{(ii)} increasing the sparsity intensity $\lambda$ leads to a sparser solution in the final map (\hyperref[t4.5]{Theorem 4.5}); and \textbf{(iii)} under higher sparsity intensity $\lambda$, the error between the biased and unbiased solutions remains controllable (\hyperref[t4.6]{Theorem 4.6}). We emphasize that these theoretical guarantees are not ends in themselves but are provided to support the practical applicability of our method. For detailed proofs of the following lemmas and theorems, please refer to \hyperref[B]{Appendix B}.

\begin{lemma} \label{l4.1}
Let $\tau$ be any sparsity penalty introduced in \S\hyperref[sec3.3]{3.3}. Assume that $P$ and $Q$ are probability measures over $\mathbb{R}^d$ with bounded support. Moreover, suppose that $\mathcal{S}(P)$ and $\mathcal{S}(Q)$ contains only L-Lipschitz continuous functions $f$ and \( g \) respectively. Then, the sparsity penalty \(\int_{\mathbb{R}^d} \tau \left( \nabla g(y) - y \right) \, dQ\) is upper bounded by $M_{\tau}$ (see \hyperref[B]{Appendix B}) for any \( g \in \mathcal{S}(Q) \).
\end{lemma}

\begin{remark} 
The bounded support assumption for $Q$ in \hyperref[l4.1]{Lemma  4.1} is reasonable, as it is feasible to impose arbitrary bounds on the parameters within the $\mathtt{ICNN}(\mathbb{R}^d)$ training space, proposed by \citet[Remark 3.4]{Makkuva20}. This assumption holds for the rest theoretical results in \S\hyperref[s4]{4}.
\end{remark}

\begin{theorem}\label{t4.3}Let \( J_{0} \) denote the Wasserstein distance as defined in equation \eqref{e4}, and let \( J_{\lambda} \) denote the biased Wasserstein distance as defined in equation \eqref{e6}. Then, we have:
\[
\lim_{\lambda \to 0} J_{\lambda} = J_{0}.
\]
\end{theorem}

\begin{theorem}\label{t4.4}Let \(\mathcal{G}\subseteq \mathcal{S}(Q)\times \mathcal{S}(P)\) be the closed set of optimal solutions 
\(\bigl(g_{0}, f_{0}\bigr)\) to the optimization problem in 
\eqref{e4}, and let \(\bigl(g_{\lambda}, f_{\lambda}\bigr)\) be an optimal solution of \eqref{e6}, 
for a given \(\lambda \ge 0\) and a convex regularizer \(\tau\), we have
\[
\lim_{\lambda \to 0} 
\mathrm{dist}\Bigl(\bigl(g_{\lambda}, f_{\lambda}\bigr),\, \mathcal{G}\Bigr) 
\;=\; 0,
\]
where
\(\mathrm{dist}\bigl((g,f), \mathcal{G}\bigr) \;=\; 
\inf_{(g_0,f_0)\in \mathcal{G}} 
\bigl\|\,(g,f)-(g_0,f_0)\,\bigr\|\)
denotes the distance from the point \(\bigl(g,f\bigr)\) to the set \(\mathcal{G}\) 
in an appropriate normed space. Note that \((g_{\lambda}, f_{\lambda})\) converges to a unique \((g_{0}, f_{0})\) only if the objective function is strongly concave-strongly convex.
\end{theorem}

\begin{theorem}\label{t4.5} Let \((g_{\lambda_{1}}, f_{\lambda_{1}})\) and \((g_{\lambda_{2}}, f_{\lambda_{2}})\) be the optimal solutions to the optimization problem defined in equation \eqref{e6} for \(\lambda = \lambda_{1}\) and \(\lambda = \lambda_{2}\), respectively, where \(0 \leq \lambda_{1} < \lambda_{2}\), and let \(\tau\) be a convex regularizers. Then, the following inequality holds:
\[
\int_{\mathbb{R}^d} \tau (\nabla g_{\lambda_{1}}(y) - y) \, dQ \geq \int_{\mathbb{R}^d} \tau (\nabla g_{\lambda_{2}}(y) - y)\, dQ.
\] 
\end{theorem}

\hyperref[t4.3]{Theorem 4.3} and \hyperref[t4.4]{4.4} establishes that the unbiasedness and convergence of the solution. \hyperref[t4.5]{Theorem 4.5} shows the monotonicity of the map's sparsity level for convex regularizers. However, we empirically observe that our non-convex regularizers also exhibit these properties, as shown in \hyperref[f5]{Figure 5}.

To align with ICNN training, we restrict our training space in (\hyperref[e6]{6}) from $\mathcal{S}(P)$ and $\mathcal{S}(Q)$ to its subspace $\mathtt{ICNN}(\mathbb{R}^d)$:
\begin{align}
\widetilde{W_2^2}(P, Q) = C_{P,Q} + &\sup_{f \in \mathtt{ICNN}(\mathbb{R}^d)} \inf_{g \in \mathtt{ICNN}(\mathbb{R}^d)}\big( \mathcal{V}_{P,Q}(f, g)  + \lambda \int_{\mathbb{R}^d} \tau \left( \nabla g(y) - y \right) \, dQ\big). \tag{7} \label{e89}
\end{align}

We also establish a bound on the deviation of the optimal transport map \(\nabla g_{\lambda}\) in (\hyperref[e6]{6}). For brevity, define  
\[
\mathcal{N}_{P,Q}(f,g) = \mathcal{V}_{P,Q}(f,g) + \lambda \int_{\mathbb{R}^d} \tau(\nabla g(y) - y) \ dQ.
\]
Let \((g_{\lambda}, f_{\lambda})\) be the optimal solution to problem~\eqref{e6} and \((g, f)\) be any candidate pair from the reparameterized space, then the optimality gaps \(\epsilon_1, \epsilon_2\) can be defined as follows:

\[
\epsilon_1(f, g) = \mathcal{N}_{P,Q}(f, g) - \inf_{g' \in \mathcal{S}(Q)} \mathcal{N}_{P,Q}(f, g'), \quad
\epsilon_2(f) = \mathcal{N}_{P,Q}(f_{\lambda}, g_{\lambda}) - \inf_{g' \in \mathcal{S}(Q)} \mathcal{N}_{P,Q}(f, g').
\]
For simplicity, let
\(s(g) = \int_{\mathbb{R}^d} \tau(\nabla g(y) - y) \, dQ,\)
where \(s(g)\) is bounded between 0 and \(M_{\tau}\) (\hyperref[l4.1]{Lemma 4.1}). Consequently, the sparsity gap between any two potentials \(g_i\) and \(g_j\), given by \(\|s(g_i) - s(g_j)\|\), is also bounded by \(M_{\tau}\). Through \(\epsilon_1\), \(\epsilon_2\), and \(M_{\tau}\), we can derive a bound on the deviation of the optimal transportation map.

\begin{theorem}\label{t4.6}
Consider \(\nabla g_{0}\) to be the unbiased ground truth map from \(Q\) to \(P\) under formulation \eqref{e4} . For $\lambda >0$ and dual potential pair \((g, f)\), where \(f\) is \(\alpha\)-strongly convex, the following bound holds for solving biased map from formulation \eqref{e6}:
\[
\|\nabla g - \nabla g_{0}\|^2_{L^2(Q)} \leq \frac{2}{\alpha} \left( \epsilon_1 + \epsilon_2 + 2\lambda M_{\tau} \right).
\]
\end{theorem}

Optimality gap is a standard technique in the OT literature for bounding map error (see~\cite[Theorem 3]{asadulaev2024neural},~\cite[Theorem 2]{fan2021scalable},~\cite[Theorem 3.6]{Makkuva20}). Given that \((g, f)\) are reparameterized from the convex space, \(\epsilon_1\) and \(\epsilon_2\) tend to converge to $0$ during training. More precisely, they can be upper bounded by the generalization error \(\delta_1\) and optimization error \(\delta_2\): \(\max\{\epsilon_1, \epsilon_2\} \le \delta_1 + \delta_2\). By using ICNNs, which have strong approximation capacity for convex dual potentials, both \(\delta_1\) and \(\delta_2\) can be controlled at very low levels. In the end, the primary source of map error then comes from the sparsity regularization.
\begin{figure*}[ht!]
    \centering
    \begin{minipage}[t]{0.62\textwidth} 
        \centering
        \includegraphics[width=0.75\textwidth]{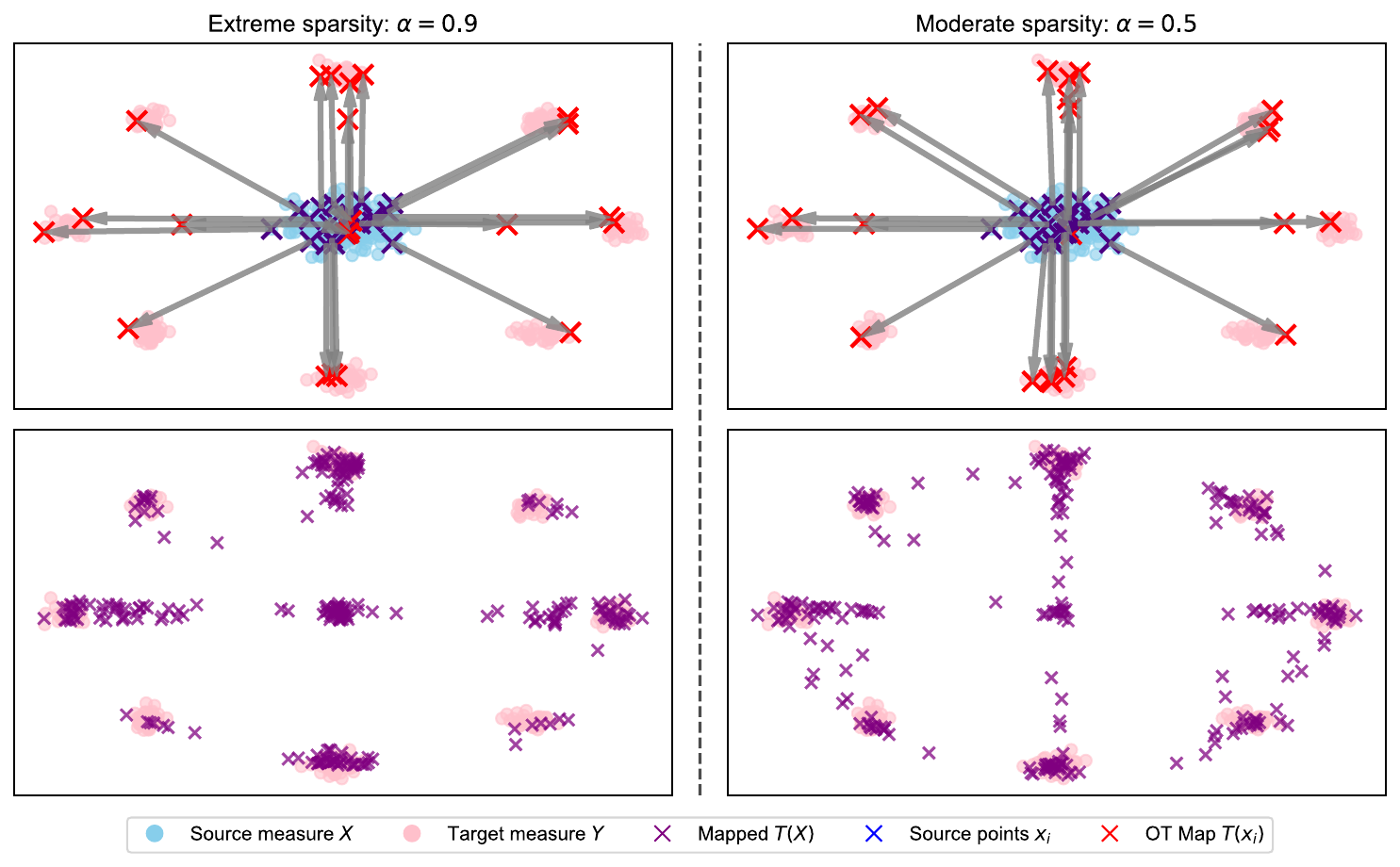} 
    \end{minipage}%
    \hfill
    \begin{minipage}[t]{0.38\textwidth} 
        \centering
        \includegraphics[width=\textwidth]{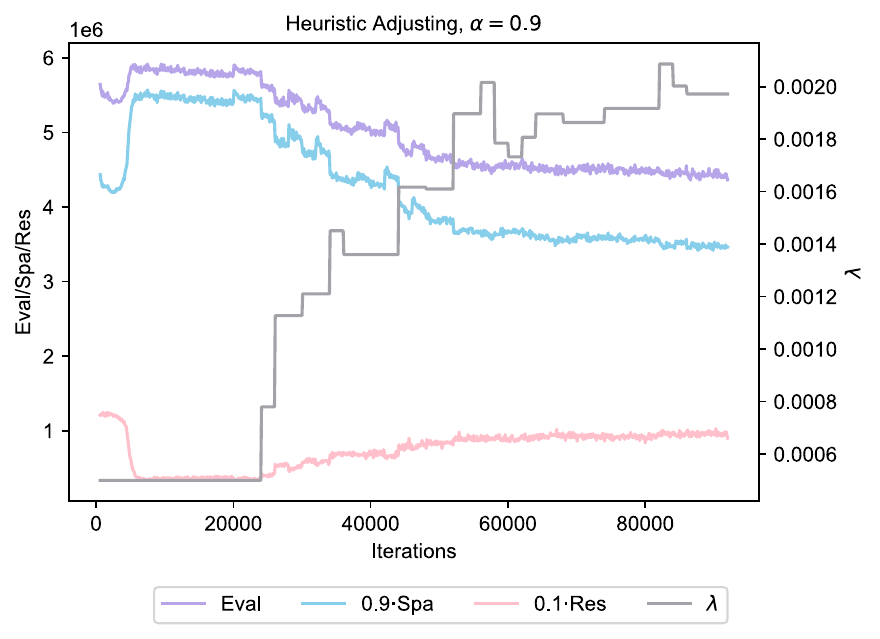} 
    \end{minipage}
    \vspace{-1em}
    \caption{\textbf{Left}: Two classic eight-Gaussian examples are presented, where the source measure is located at the center, and eight target measures are generated from Gaussian distributions. These examples illustrate the trade-off between the sparsity of the map and its feasibility under varying values of $a$. \textbf{Right}: A simulation example demonstrates how $\lambda$ changes during simulated annealing training under $a=0.9$, along with the corresponding levels of map sparsity and feasibility.}
    \label{f4}
\end{figure*}

\subsection{Practical Scheme with Heuristic Control}

In practice, inducing displacement sparsity in neural OT solvers are faced the following challenges: 

\textbf{(i)} Same as exact OT sovlers, if \(\lambda\) is set too low, the resulting map lacks sufficient sparsity. Conversely, if \(\lambda\) is set too high, the accuracy of the map degrades, leading to a larger deviation of \(T_{\#}(P)\) from the target measure \(Q\). This trade-off is clearly illustrated in \hyperref[f2]{Figure 2} and additional example in \hyperref[f4]{Figure 4} via classic eight-Gaussian setup. Note that eight-Gaussian example is intended solely for illustrative purposes, as sparsity is not a necessary requirement in this context. In the extreme sparsity case, more and more points collapse towards the source center without adequate transportation to their targets.

\textbf{(ii)} Specifically, in large-scale, high-dimensional neural training, optimization under a static objective is prone to early-stage convergence to suboptimal solutions, resulting in transport maps that deviate from the ground-truth low-dimensional solution. Our experiments demonstrate that using a adaptive objective—by gradually adjusting the sparsity intensity \(\lambda\) during training—yields solutions with lower dimension and better match the true dimensionality compared to the constant-\(\lambda\) objective.

\textbf{Rationale.} Both challenges indicate the necessity of designing a adaptive framework to adjust the sparsity intensity \(\lambda\). This is particularly beneficial for end-users conducting downstream analysis on high-dimensional data, as an automatically adjusted \(\lambda\) eliminates the need for manual tuning and yields more interpretable transport maps with even lower effective dimensionality. To this end, we propose a heuristic framework tailored to both low- and high-dimensional tasks. For low-dimensional settings, where the target dimensionality is often ambiguous, we allow users to navigate the trade-off between map's feasibility and sparsity, selecting the final map through a result-driven tuning process. In contrast, for high-dimensional tasks, we directly constrain the map to match the user’s desired dimensionality, while searching for the most feasible solution under this constraint.

\begin{algorithm}[h]
\caption{Simulated‑Annealing Heuristic for Adaptive Regularization $\lambda$}
\label{a1}
\begin{algorithmic}[1]

\REQUIRE Initial weight $\lambda_0$, Initial temperature $\text{Tem}_0$, minimum temperature $\text{Tem}_{\min}$, decay $d$
\REQUIRE Initiation steps $n_{\text{ini}}$, train steps $n_{\text{tr}}$, rollback steps $n_{\text{sm}}$; Tradeoff weight $a$

\STATE \textbf{Initialization:} train $(f,g)$ with $\lambda_0$ for $n_{\text{ini}}$ iterations
\STATE $\text{Tem} \gets \text{Tem}_0$, $\lambda \gets \lambda_0$

\WHILE{$\text{Tem} > \text{Tem}_{\min}$}
    \STATE $\lambda' \gets \textproc{SimulatedAnnealingProposal}(\text{Tem}, \lambda)$ \hfill \texttt{\# Generating new $\lambda$}
    \STATE Train $(f,g)$ with $\lambda'$ for $n_{\text{tr}}$ iterations and obtain map $T_n$
    \STATE $\text{Eval} \gets \textproc{Evaluate} (T_n,a)$ \hfill \texttt{\# Getting evaluation metric for map $T_{n}$}       
    \IF{$\textproc{Accept} (\text{Eval},\text{Tem})$}
        \STATE $\lambda \gets \lambda'$ \hfill \texttt{\# Metropolis‑style acceptance to encourage exploration}           
    \ELSE
        \STATE Train $(f,g)$ with current $\lambda$ for $n_{\text{sm}}$ iterations
    \ENDIF
    \STATE $\text{Tem} \gets d \cdot \text{Tem}$   
\ENDWHILE
\end{algorithmic}
\end{algorithm}

\textbf{Low-Dimensional Tasks.} We introduce a parameter $\alpha$ to tune the tradeoff by adaptively adjusting \(\lambda\) to optimize a goal function that balances sparsity and feasibility during the ICNN training process. For the map \(T_{n}: X\rightarrow Y\) learned at the \(n\)-th iteration of the ICNN training, we define its sparsity and feasibility level through two metrics, Spa (Sparsity) and Res (Residue), where
\[\text{Spa} = \int_P \tau(T_n(x) - x) \, dP, \quad \text{Res}=D(T_{n\#}(P), Q).\]
We use the corresponding sparsity metric, $\tau$, during training, while measuring the error through the divergence $D$ between the mapped distribution and the actual target distribution. We approximate the source and target distributions, $P$ and $Q$, as finite sets of sampled points, and use the Wasserstein distance to quantify such geometric divergence. We accelerate the training process by approximating it with the sliced Wasserstein distance \cite{bonneel2015sliced}. Eventually, we construct the evaluation function:
\[
\label{e8}
\text{Eval} = a \cdot \text{Spa} + (1 - a) \cdot \text{Res}, \quad a \in [0,1]. \tag{8}
\]
The objective is to find the value of $\lambda$ that minimizes $\text{Eval}$. Since $\lambda$ is not explicitly related to $\text{Eval}$, we embed a simulated annealing search framework into the ICNN. A detailed technical walkthrough of \hyperref[a1]{Algorithm~1} is provided in \hyperref[ac]{Appendix C}, and the simulation trajectory is shown in \hyperref[f4]{Figure~4} (Right).

\textbf{High-Dimensional Tasks.} While low-dimensional tasks are usually applied to synthetic data for simulation purposes, high-dimensional tasks arise in more practical settings such as single-cell RNA (scRNA) and drug perturbation studies, where the goal is to match cells and track changes in high-dimensional gene or cell features. Learning a transport map with reduced displacement dimensionality, under a smaller set of genes or features, can greatly aid downstream analysis.

We adapt the heuristic search framework from \hyperref[a1]{Algorithm~1}, removing weight $a$ from \eqref{e8} to impose a direct dimensionality constraint $l$. If $l$ is lower than solver's achievable dimension, the solver is pushed toward the lowest possible dimension; if $l$ is attainable, the solver finds the most feasible transport map that meets constraint $l$. Technical details are provided in \hyperref[a4]{Algorithm 4} (\hyperref[ac]{Appendix C}). Our experiments further demonstrate the necessity of adaptive-$\lambda$ control for high-dimensional tasks.

\section{Experiments}

\textbf{Design \& Setup.} We contextualize our method for drug perturbation tasks, aligning with CellOT \cite{Bunne23}. In this setting, OT is used to match perturbed cells to their original state based on high-dimensional features such as scRNA expression and 4i (Iterative Indirect Immunofluorescence Imaging) data. The baseline in CellOT is ICNN OT \cite{Makkuva20}; we also compare our method to the exact OT solver from \cite{Cuturi23}, which provides higher accuracy but is less practical for its computation and memory inefficiency.

\subsection{Synthesized sc-RNA Perturbation Task} 
To evaluate how well our method can recover the ground-truth displacement dimensionality, we began with a synthetic dataset instead of real data, which lack reliably known ground truth.\begin{wrapfigure}{r}{0.39\textwidth}
  \centering
\includegraphics[width=0.38\columnwidth,height=0.16\textheight]{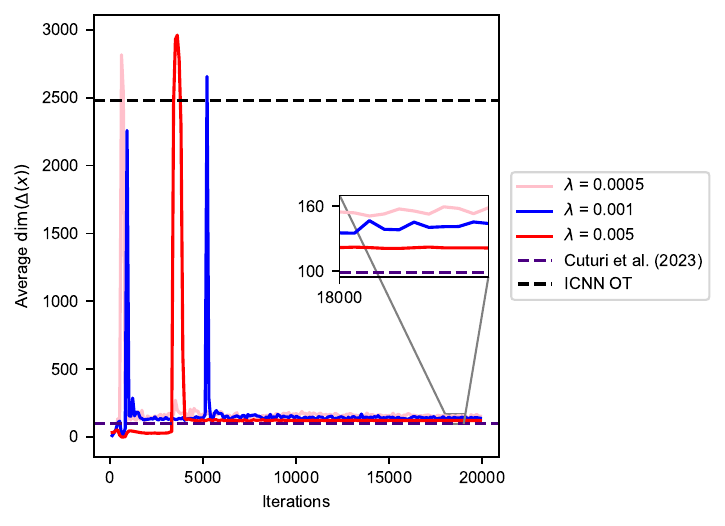} 
    \vspace{-1em}
    \caption{Synthesized sc-RNA perturbation dataset with \( n=4000 \), \( d=3000 \), and \( k=100 \). The average displacement dimensionality is shown.}
    \label{f5}
    \vspace{-2em}
\end{wrapfigure}  We used the exact OT solver ($\ell_{1}$) to help construct the synthetic dataset, ensuring that the ground-truth displacement dimensionality is at least attainable. If the exact solver fails to reach it, the dataset design would be considered flawed. 

The dataset is constructed as follows: we first define \( n \), the number of cells before and after drug perturbation; \( d \), the total number of genes per cell (overall dimensionality); and \( k \), the number of genes truly affected by the drug (ground-truth dimensionality). Random noise is added to the remaining genes to simulate real noise. The controlled cells are treated as the source, and the perturbed cells are treated as the target. The objective is to evaluate whether these methods can provide a mapping that focuses exclusively on the truly perturbed genes while disregarding the noisy ones. Example result is shown in \hyperref[f5]{Figure 5} and \hyperref[f10]{Figure 10}, where \hyperref[f10]{Figure 10} also includes further comparison of effectiveness between $\ell_{1}$ and $\ell_{0}$-norm in the neural OT setting.

\begin{wraptable}{t}{0.6\textwidth}
\captionof{table}{Performance on synthesized task.}
\label{tab:ot_comparison}
\centering
\resizebox{0.59\textwidth}{!}{%
\begin{tabular}{lcc|ccccc}
\toprule
\textbf{Method} & \textbf{ICNN OT} & \textbf{Cuturi et al. 23} & \textbf{Ours} & $\lambda = 5 \times 10^{-4}$ & $\lambda = 1 \times 10^{-3}$ & $\lambda = 5 \times 10^{-3}$ & adaptive-$\lambda$ \\
\midrule
\textbf{dim($\Delta(\mathbf{x})$)} & 2480 & 99 & & 153 & 138 & 120 & 107 \\
\bottomrule
\end{tabular}%
\label{t1}
}
\vspace{-1em}
\end{wraptable}

\textbf{Adaptive-$\lambda$ Achieves Better Results.} In \hyperref[t1]{Table~1}, we present the results of our method using both constant-$\lambda$ and adaptive-$\lambda$ methods. First, the ground-truth displacement dimensionality is reachable by the exact OT solver from \citet{Cuturi23}, validating the design of our synthetic dataset. Moreover, adaptive-$\lambda$ outperforms the constant-$\lambda$ settings by achieving a displacement dimensionality closer to the ground truth. This underscores the importance of using an adaptive objective in neural OT training to avoid suboptimal solutions. To further support the robustness of adaptive-$\lambda$, we include training trajectories across multiple runs and more results for \hyperref[t1]{Table 1} in \hyperref[ad1]{Appendix D.1}.

\textbf{Chosen Genes Accuracy.} On multiple synthetic datasets, we also evaluate how many of the selected genes overlap with the ground-truth perturbed genes. Our method achieves accuracies of over 95\% and 99\% on the $k=100,500$ datasets, respectively. Detailed results are provided in \hyperref[ad2]{Appendix D.2}.
\vspace{-0.5em}
\subsection{Real 4i Perturbation Task} \label{sec5.2}
\begin{figure*}[ht!]
\vspace{-0.5em}
    \centering
    \includegraphics[width=0.85\textwidth]{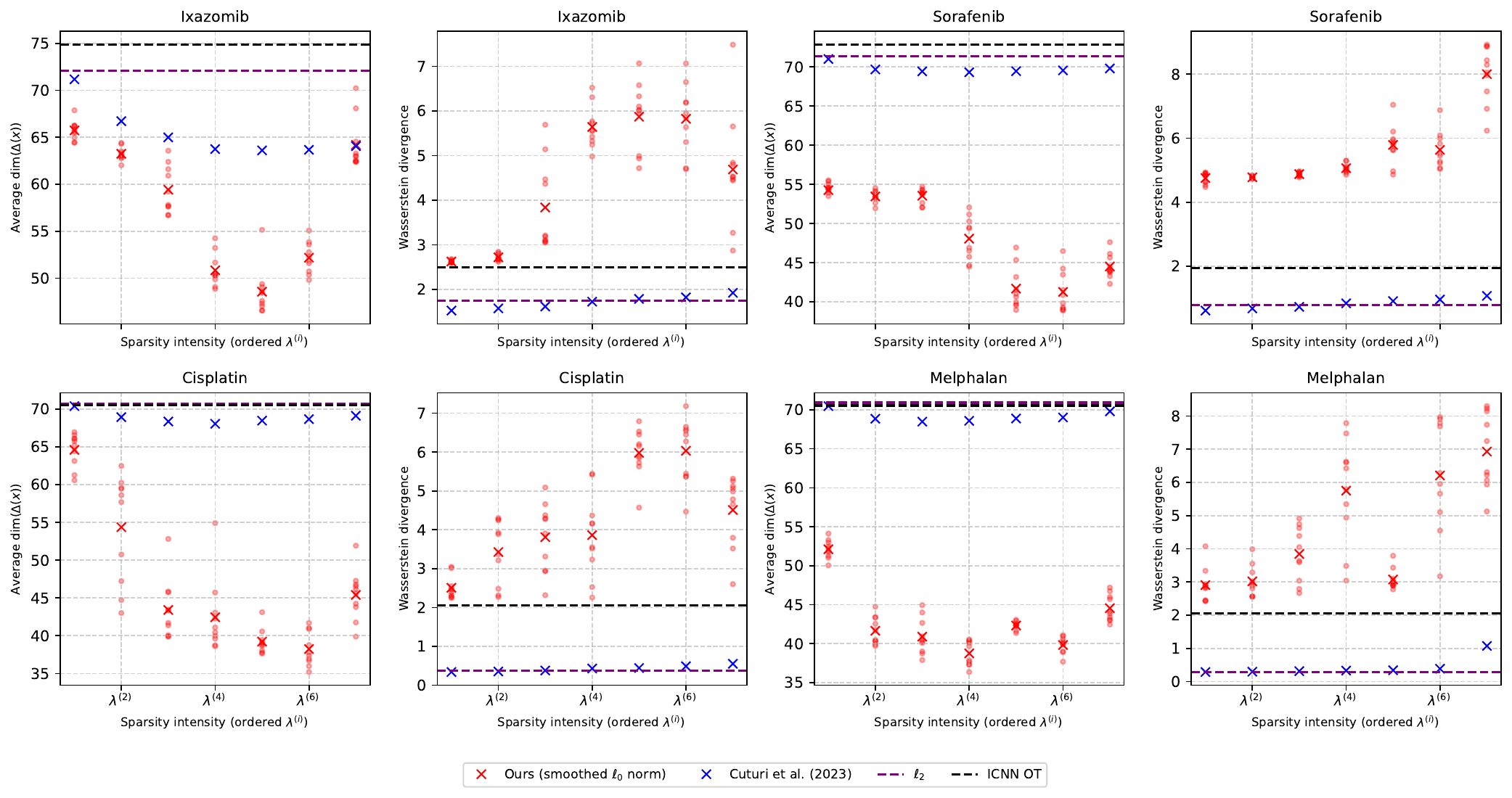}
    \vspace{-0.5em}
    \caption{The results of the 4i perturbation are presented. The average of ten runs is shown for each drug. $\lambda^{(i)}$ represents the ordered statistics of each set of $\lambda$ used in the experiments.}
    \label{f6}
    \vspace{-1em}
\end{figure*}

We use the preprocessed 4i cell perturbation dataset from CellOT \cite{Bunne23}. For details on the experimental and evaluation setup, please refer to \hyperref[ad3]{Appendix D.3}. We report two metrics: (\textbf{i}) the average dimensionality of the displacement vectors—\textit{lower values indicate better interpretability}; and (\textbf{ii}) the Wasserstein divergence between $T_{\#}(P)$ and $Q$—\textit{lower values indicate better feasibility}. We conduct four experiments across different drugs to demonstrate the superiority of our method on real datasets.

\textbf{$\ell_{0}$ v.s. $\ell_{1}$ Norm.} Notably, our $\ell_0$ norm regularization outperforms the $\ell_1$-based exact OT solver from \citet{Cuturi23} in this specific setting. This is particularly valuable for real datasets where perturbed and noise genes are not clearly separable by magnitude ($\ell_{1}$-norm), making direct dimensionality regularization ($\ell_{0}$-norm) more effective. Although the exact OT solver may achieve higher accuracy when the magnitude gap is clear, our method proves more robust under ambiguity. Furthermore, we show that our $\ell_0$ norm outperforms the $\ell_1$ norm under the neural OT solver (see \hyperref[ad3]{Appendix D.3} for details). These results highlight the strength of our biased minimax formulation, which better accommodates general elastic costs and more effectively enforces low-dimensional structure.

\vspace{-0.5em}
\section{Conclusion}

We present more experimental results and analysis along with a discussion of limitations in \hyperref[ad]{Appendix D}. Overall, we propose a formulation for learning displacement-sparse neural OT maps and demonstrate its effectiveness through both theoretical analysis and empirical results. Our novel $\ell_0$ regularizer is well aligned with this formulation and outperforms existing alternatives. To support practical usage, we also provide an adaptive framework that enables efficient and robust deployment. We hope future work will build upon our formulation to explore further sparse or interpretable structures in neural OT solvers.



\bibliographystyle{plainnat}
\bibliography{ref}


\newpage
\appendix
\section{Further Related Works}\label{a1}

\textbf{Neural OT with General Cost.} In this section, we further discuss the limitations associated with neural OT methods using general cost functions. It is important to clarify that many neural OT solvers are designed to approximate the Kantorovich coupling $\Pi$ rather than the Monge map $T$. In contrast, our work aims to recover the Monge map $T$, and it is generally infeasible to exactly convert a Kantorovich coupling $\Pi$ into a Monge map $T$.

\cite{asadulaev2024neural} requires the objective function to be strictly convex $w.r.t.$ to the coupling $\Pi$ to guarantee that the solution obtained by the minimax formulation is indeed an OT map \cite[Theorem 2]{asadulaev2024neural}. But this may necessitate regularization such as negative entropy which leads to further deviation from the OT map and potential numerical instability issues. \cite{uscidda2023monge} proposes a relatively conceptual approach by introducing the Monge gap as regularization. For minimization of Monge map, usually its gradient information is required. This necessitates solving another OT problem in each iteration to estimate the gradient (see \cite[\S4.1]{uscidda2023monge}). \cite{fan2021scalable} directly requires second-order differentiability (see \cite[Assumption 1]{fan2021scalable}). \cite{buzun2024enot} requires strictly convex cost $h$ and has barely any theoretical guarantees for the framework proposed and the expectile regression regularization may not always fulfill the mission to facilitate the $c$-concavity criterion in \citet{villani2009optimal}'s book.

\textbf{Further Sparsity-related Optimal Transport.} Apart from displacement sparsity, other works have explored different aspects of sparsity in the context of OT. \citet{pmlr-v139-fan21d} pioneer the use of strongly convex regularizers, including $\ell_{2}$ and group-$\ell_{1}$ (lasso) penalties, to obtain smooth and sparse OT solutions. This approach yields sparse and group-sparse transport plans while controlling approximation error. \citet{liu2023sparsity} introduced a regularization method to control the column cardinality of the transportation plan matrix, enhancing both storage efficiency and computational complexity. Similarly, \citet{blondel18} proposed a regularization approach using the \(\ell^2_2\)-norm to address the dense and complex transportation plans typically produced by the entropic-regularized Sinkhorn solution. \citet{luo2023group} develop Group Sparse OT for a process flexibility design problem in manufacturing, which could solve multiple OT problems jointly with a group-lasso regularizer that induces a consistent sparse support across all transport plans. \citet{terjek2022optimal} extend OT theory to general $f$-divergence regularizers, recovering known cases (entropy, quadratic) and identifying divergences that yield naturally sparse solutions. \citet{Bao2022Sparse} propose a deformed $q$-entropy regularization that bridges entropy ($q\to 1$) and quadratic regularization ($q\to 0$), showing that as $q\to 0$ the OT plan becomes increasingly sparse, offering a tunable trade-off between sparsity and convergence speed. \citet{gonzalez2024sparsity} provide a precise analysis of the sparsity of quadratically-regularized OT in the continuous 1D case, which proves that as the regularization parameter $\varepsilon\to 0$, the support of the optimal plan converges to the sparse Monge solution at a quantified rate. \citet{nutz2024quadratically} investigates existence and uniqueness of dual potentials in quadratically-regularized OT, showing that non-uniqueness of dual potentials is directly tied to sparsity of the primal optimal plan. \citet{zhang2023manifold} use quadratically-regularized OT in a symmetric form for manifold learning. The quadratic penalty yields a sparse affinity matrix that adapts to data, effectively selecting a sparse subset of neighbors for each point. \citet{andrade2024sparsistency} studies an inverse OT problem, inferring the ground cost from observed couplings, with an $\ell_{1}$ regularizer to encourage a cost with sparse feature interactions. 

\textbf{Wasserstein Generative Adversarial Network.} It is also worth clarifying that approaches based on the Wasserstein Generative Adversarial Network (W-GAN)~\cite{wgan,gulrajani2017improved,wei2018improving} do not compute OT mappings or transportation plans directly; rather, they use the Wasserstein distance as a loss function during training.

\section{Proofs of Theorems}
\label{B}

\noindent \textbf{Proof of \hyperref[l4.1]{Lemma 4.1}:} Given the assumption on the bounded support of the target measure and the boundedness of $\nabla g$, which is implied by the uniform \emph{Lipschitiz} continuity of $g$ in $\mathcal{S}(Q)$ and convexity, we can to show that \(\tau(\cdot)\) is bounded:
\begin{itemize}
  \item The smoothed \(\ell_{0}\)-norm is bounded because the dimensionality of the target measures is bounded.
  \item The \(\ell_{1}\)-norm is bounded because the magnitude of the displacement vectors is bounded.
  \item By the Weierstrass theorem, \(\tau_{\mathrm{stvs}}\) is bounded on the compact region covering the value of $\nabla g(y)-y$ since it is continuous.
\end{itemize}
Given these bounds, we can conclude that \hyperref[l4.1]{Lemma  4.1} holds for all three sparsity penalties used in our analysis with upper bound $M_{\tau}$.

\begin{theorem}[von Neumann's Minimax Theorem] \label{minimax}
Let \( X \subset \mathbb{R}^n \) and \( Y \subset \mathbb{R}^m \) be two compact and convex sets. Suppose \( f: X \times Y \to \mathbb{R} \) is a continuous function that is concave in \( x \) for each fixed \( y \) and convex in \( y \) for each fixed \( x \) (concave-convex). Then, the following equality holds:
\[
\max_{x \in X} \min_{y \in Y} f(x, y) = \min_{y \in Y} \max_{x \in X} f(x, y).
\]
\noindent Moreover, assuming the optimal solution $(x^*, y^*)$ exists, then it satisfies the following saddle point inequality:
\[
f(x,y^*)\leq f(x^*, y^*) \leq f(x^*, y),\quad \forall x \in X, y \in Y,\
\]
\end{theorem}
In the context of ICNN, our sets $X$ and $Y$ are instead topological vector spaces, which places them under the scope of a variant of von Neumann’s Minimax Theorem: Sion’s Minimax Theorem \cite{sion}.
\begin{lemma}\label{lA2}
For equation \eqref{e4}, given \( f \in \mathtt{CVX}(P) \) and \( g \in \mathtt{CVX}(Q) \), the function
\[
\mathcal{V}_{P,Q}(f, g) = -\mathbb{E}_P[f(X)] - \mathbb{E}_Q\left[\langle Y, \nabla g(Y)\rangle - f(\nabla g(Y))\right]
\]
is a concave-convex and continuous function.
\end{lemma}

\noindent \textbf{Proof of Lemma A.2 --- \emph{Concave-Convex}:} To prove that \( \mathcal{V}_{P,Q}(f, g) \) is a concave-convex function, we first show that, for a fixed \( g \), \( \mathcal{V}_{P,Q}(f, g) \) is concave in \( f \).

Consider any two functions \( f_{1}, f_{2} \in \mathtt{CVX}(P) \) and any \( \alpha \in [0,1] \). We aim to show that:
\[
\mathcal{V}_{P,Q}(\alpha f_{1} + (1-\alpha) f_{2}, g) \geq \alpha \mathcal{V}_{P,Q}(f_{1}, g) + (1-\alpha) \mathcal{V}_{P,Q}(f_{2}, g). \tag{9} \label{e9}
\]
To demonstrate this, note that by linearity:
\begin{align*}
\mathcal{V}_{P,Q}(\alpha f_{1} + (1-\alpha) f_{2}, g) &= -\mathbb{E}_P[(\alpha f_{1} + (1-\alpha) f_{2})(X)] \\
&\quad - \mathbb{E}_Q\left[\langle Y, \nabla g(Y)\rangle - (\alpha f_{1} + (1-\alpha) f_{2})(\nabla g(Y))\right] \\
&= - \alpha \mathbb{E}_P[f_{1}(X)] - (1-\alpha) \mathbb{E}_P[f_{2}(X)] \\
&\quad + \alpha \mathbb{E}_Q\left[f_{1}(\nabla g(Y))\right] + (1-\alpha) \mathbb{E}_Q\left[f_{2}(\nabla g(Y))\right] \\
&\quad - \mathbb{E}_Q\left[\langle Y, \nabla g(Y)\rangle\right] \\
&= \alpha \mathcal{V}_{P,Q}(f_{1}, g) + (1-\alpha)\mathcal{V}_{P,Q}(f_{2}, g).
\end{align*}
This establishes that \( \mathcal{V}_{P,Q}(f, g) \) is concave in \( f \) for a fixed \( g \).

Next, we show that for a fixed \( f \), \( \mathcal{V}_{P,Q}(f, g) \) is convex in \( g \). Consider any \( g_{1}, g_{2} \in \mathtt{CVX}(Q) \) and any \( \alpha \in [0,1] \). We need to prove that:
\[
\mathcal{V}_{P,Q}(f, \alpha g_{1} + (1-\alpha) g_{2}) \leq \alpha \mathcal{V}_{P,Q}(f, g_{1}) + (1-\alpha)\mathcal{V}_{P,Q}(f, g_{2}). \tag{10} \label{e10}
\]
By definition, we have:
\begin{align*}
\mathcal{V}_{P,Q}(f, \alpha g_{1} + (1-\alpha) g_{2}) 
&= -\mathbb{E}_P[f(X)] \\
&\quad - \mathbb{E}_Q\left[\langle Y, \alpha \nabla g_{1} (Y) + (1-\alpha) \nabla g_{2}(Y)\rangle - f\big(\alpha \nabla g_{1} (Y) + (1-\alpha) \nabla g_{2}(Y)\big)\right]  \\
&= - \mathbb{E}_P[f(X)] - \mathbb{E}_Q\left[\langle Y, \alpha \nabla g_{1} (Y)\rangle + \langle Y, (1-\alpha) \nabla g_{2} (Y)\rangle\right] \\
&\quad +   \mathbb{E}_Q\left[f\big(\alpha \nabla g_{1} (Y) + (1-\alpha) \nabla g_{2}(Y)\big)\right] \\
&\leq -\mathbb{E}_P[f(X)] - \mathbb{E}_Q\left[\langle Y, \alpha \nabla g_{1} (Y)\rangle + \langle Y, (1-\alpha) \nabla g_{2} (Y)\rangle \right] \\
&\quad  + \mathbb{E}_Q\left[\alpha f( \nabla g_{1} (Y)) + (1-\alpha)f( \nabla g_{2}(Y))\right] \quad (\because \text{$f$ is convex}).
\end{align*}

Combining the terms, we obtain the desired inequality (\hyperref[e10]{10}). Therefore, by combining inequalities (\hyperref[e9]{9}) and (\hyperref[e10]{10}), we conclude that \( \mathcal{V}_{P,Q}(f, g) \) is indeed a concave-convex function. \hfill $\square$

\noindent \textbf{Proof of Lemma A.2 --- \emph{Continuity}:} We aim to prove that $\mathcal{V}_{P,Q}(f,g)$ is continuous under the assumptions: \textbf{(i)} \(Q\) has \emph{bounded support}; \textbf{(ii)} Every \(g\) in \(\mathcal{S}(Q)\) is \(L\)-\emph{Lipschitz}, which means \(\|\nabla g(\cdot)\|\le L\). We assume that $\nabla g(\cdot)$ is \(L\)-\emph{Lipschitz}; \textbf{(iii)} Every \(f\) in \(\mathcal{S}(P)\) is also \(L\)-\emph{Lipschitz}; \textbf{(iv)}  We assume \(\mathcal{S}(P)\) and \(\mathcal{S}(Q)\) under the \emph{uniform convergence} topology, i.e., \((f_n,g_n)\to(f,g)\) means \(f_n\to f\) uniformly and \(g_n\to g\) uniformly on their respective domains. Moreover, given $f$ and $g$ is \(L\)-\emph{Lipschitz} and convex, it's sufficient to show that $(f_n, g_n) \rightarrow (f,g)$ is also under epi-convergence.

By \citet[Theorem~12.35]{rockafellar1998variational}, if \(g_n\) are convex and epi-convergent to a convex \(g\), then under suitable regularity conditions (each \(g_n\) being differentiable on a compact domain, with uniformly bounded gradients), we get \(\nabla g_n(y) \to \nabla g(y)\) \emph{pointwise in \(y\)}.  

Since each \(g_n\) is \(L\)-\emph{Lipschitz} on a compact domain, it follows that \(\|\nabla g_n(\cdot)\|\le L\).  Moreover, the family \(\{\nabla g_n\}\) is \emph{equicontinuous}:
\[
  \|\nabla g_n(x) - \nabla g_n(y)\|\leq \|\nabla g_n(x)\|+\|\nabla g_n(y)\|\leq2L
\]
Under Arzelà–Ascoli Theorem, for a compact set, \emph{equicontinuity} plus \emph{pointwise convergence} imply \(\nabla g_n\to \nabla g\) \emph{uniformly}. Therefore, we have
\[
  \nabla g_n(y)
  \;\longrightarrow\;
  \nabla g(y)
  \quad
  \text{uniformly for }y\in \mathrm{supp}(Q).
\]
Since each \(f_n\) is also \(L\)-\emph{Lipschitz}, we have uniform continuity on any bounded set.  
Because \(\|\nabla g_n(y)\|\le L\), \(\nabla g_n(y)\) indeed lies in the ball of radius \(L\), which we denote this ball by \(\mathcal{B}_L(0)\).  

For \(y\) in the bounded support of \(Q\), we then examine
\[
   \bigl|\, f_n\bigl(\nabla g_n(y)\bigr)
           \;-\; f\bigl(\nabla g(y)\bigr)\bigr|
   \;\le\;
   \underbrace{\bigl|\,f_n\bigl(\nabla g_n(y)\bigr)
               - f\bigl(\nabla g_n(y)\bigr)\bigr|}_{\text{(I)}}
   \;+\;
   \underbrace{\bigl|\,f\bigl(\nabla g_n(y)\bigr)
               - f\bigl(\nabla g(y)\bigr)\bigr|}_{\text{(II)}}.
\]
For part (I), since \(f_n\to f\) in uniform convergence and bounded by uniform norm, we get
\[
  \sup_{\|z\|\le L}\,\bigl|\,f_n(z)-f(z)\bigr|
  \;\xrightarrow[n\to\infty]{}\;0.
\]
Therefore, \(\bigl|f_n(\nabla g_n(y)) - f(\nabla g_n(y))\bigr|\to 0\) uniformly in \(y\).

For part (II), because \(f\) is \(L\)-\emph{Lipschitz}, it's then uniformly continuous on the ball \(\mathcal{B}_L(0)\).  Hence if \(\|\nabla g_n(y)- \nabla g(y)\|\le \delta\), then we have
\[
   \bigl|\,f(\nabla g_n(y)) - f(\nabla g(y))\bigr|
   \;<\;\varepsilon(\delta),
\]
where \(\varepsilon(\delta)\to0\) as \(\delta\to 0\).  
Since \(\nabla g_n\to \nabla g\) uniformly, \(\|\nabla g_n(y)-\nabla g(y)\|\) can be made small independently of \(y\).  

Integrating (I) and (II) together, we can conclude
\[
   f_n\bigl(\nabla g_n(y)\bigr)
   \;\longrightarrow\;
   f\bigl(\nabla g(y)\bigr)
   \quad \text{uniformly in }y\in\mathrm{supp}(Q).
\]

After studying the behavior of $f_{n}(\nabla g_{n}(y))$, we finally move the whole part of $\mathcal{V}_{P,Q}(f,g)$; we first define
\[
   \mathcal{V}_{P,Q}(f_n,g_n)
   \;=\; 
   -\mathbb{E}_P\bigl[f_n(X)\bigr]
   \;-\;
   \mathbb{E}_Q\Bigl[\langle Y,\nabla g_n(Y)\rangle 
                     \;-\; f_n\bigl(\nabla g_n(Y)\bigr)\Bigr].
\]
We then compare \(\mathcal{V}_{P,Q}(f_n,g_n)\) and \(\mathcal{V}_{P,Q}(f,g)\) part by part: 
\begin{align*}
\bigl|\mathcal{V}_{P,Q}(f_n,g_n) - \mathcal{V}_{P,Q}(f,g)\bigr|
&\le
\underbrace{\bigl|\mathbb{E}_{P}[f_n(X)] - \mathbb{E}_{P}[f(X)]\bigr|}_{\text{(I)}} \\
& +\;
\underbrace{\Bigl|\mathbb{E}_{Q}\Bigl[\langle Y,\nabla g_n(Y)\rangle - f_n\bigl(\nabla g_n(Y)\bigr)\Bigr]
-
\mathbb{E}_{Q}\Bigl[\langle Y,\nabla g(Y)\rangle - f\bigl(\nabla g(Y)\bigr)\Bigr]
\Bigr|}_{\text{(II)}}.
\end{align*}

For part (I), since \(X\) lies in the bounded support of \(\mathcal{S}(P)\) and \(f_n\to f\) uniformly, we have \(f_n(X)\to f(X)\) for all \(X\).  Uniform convergence on a bounded support measure implies 
\(\mathbb{E}_P[f_n(X)]\to \mathbb{E}_P[f(X)]\).

For part (II), we have already showed that \(\nabla g_n(Y)\to \nabla g(Y)\) and \(f_n(\nabla g_n(Y))\to f(\nabla g(Y))\) \emph{uniformly} in \(Y\).  Also, \(\langle Y,\nabla g_n(Y)\rangle \to \langle Y,\nabla g(Y)\rangle\) uniformly because \(Y\) lies in a bounded set and \(\nabla g_n(Y)\) is uniformly bounded and convergent to \(\nabla g(Y)\).
Therefore,
\(\bigl[\langle Y,\nabla g_n(Y)\rangle - f_n(\nabla g_n(Y))\bigr]\)
converges uniformly to 
\(\bigl[\langle Y,\nabla g(Y)\rangle - f(\nabla g(Y))\bigr]\).  
Since \(Q\) is a probability measure on a bounded domain and the integrand is uniformly bounded (by a constant depending on \(L\) and the radius of \(\mathcal{S}(Q)\)), we can again use standard dominated‐convergence arguments to conclude
\[
  \mathbb{E}_{Q}\Bigl[\langle Y,\nabla g_n(Y)\rangle - f_n(\nabla g_n(Y))\Bigr]
  \;\longrightarrow\;
  \mathbb{E}_{Q}\Bigl[\langle Y,\nabla g(Y)\rangle - f(\nabla g(Y))\Bigr].
\]
Combining (a) and (b), we see that
\[
   \lim_{n\to\infty}\,\bigl|\mathcal{V}_{P,Q}(f_n,g_n) 
     - \mathcal{V}_{P,Q}(f,g)\bigr|
   \;=\;0,
\]
which shows
\begin{align*}
  \mathcal{V}_{P,Q}(f_n,g_n)
  \;\longrightarrow\;
  \mathcal{V}_{P,Q}(f,g).
\end{align*}
Therefore, \(\mathcal{V}_{P,Q}\) is continuous with respect to \((f,g)\) in the uniform convergence topology under the stated assumptions.

With \hyperref[l4.1]{Lemma 4.1}, Theorem A.1, and Lemma A.2, we can then prove \hyperref[t4.3]{Theorem 4.3}. \hfill $\square$

\textbf{Proof of \hyperref[t4.3]{Theorem 4.3}:} As defined in \S\hyperref[s4]{4}, let \( s(g) = \int_{\mathbb{R}^d} \tau (\nabla g(y) - y) \, dQ \). We first aim to prove that \( J_{\lambda} \geq J_{0} \). This follows from the non-negativity of \( \tau(\cdot) \) and \( \lambda \), which implies:
\[
\mathcal{V}_{P,Q}(f, g) + \lambda s(g) \geq \mathcal{V}_{P,Q}(f, g).
\]
Taking the infimum over \( g \) on both sides and then the supremum over \( f \), we obtain:
\[
\sup_{f \in \mathtt{ICNN}(\mathbb{R}^d)} \inf_{g \in\mathtt{ICNN}(\mathbb{R}^d)}  \left( \mathcal{V}_{P,Q}(f, g) + \lambda s(g) \right) \geq \sup_{f \in \mathtt{ICNN}(\mathbb{R}^d)} \inf_{g \in \mathtt{ICNN}(\mathbb{R}^d)}  \mathcal{V}_{P,Q}(f, g),
\]
which is equivalent to the inequality \( J_{\lambda} \geq J_{0} \). To show that \( \lim_{\lambda \to 0} J_{\lambda} = J_{0} \), it remains to prove that \( \lim_{\lambda \to 0} J_{\lambda} \leq J_{0} \).

Let \( (f_{0}, g_{0}) \) be the optimal solution corresponding to \( J_{0} \). Since \( f \) and \( g \) are parameterized from a convex space and \( \mathcal{V}_{P,Q}(f, g) \) is a concave-convex function (as established in Lemma A.3), by Theorem A.1, we have:
\[
\mathcal{V}_{P,Q}(f, g_{0}) \leq \mathcal{V}_{P,Q}(f_{0}, g_{0}) = J_{0}. \tag{11} \label{e11}
\]
Now, consider the compactness of the function space \(\mathtt{ICNN}(\mathbb{R}^d)\), there exists \( (\widetilde{f_{\lambda}}, \widetilde{g_{\lambda}}) \) that is the optimal solution corresponding to \( J_{\lambda} \). For this supremum-infimum formulation, we have:
\[
J_{\lambda} = \mathcal{V}_{P,Q}(f_{\lambda}, g_{\lambda}) + \lambda s(g_{\lambda}) \leq \mathcal{V}_{P,Q}(f_{\lambda}, g) + \lambda s(g). \tag{12} \label{e12}
\]
Note that Theorem A.1 does not necessarily apply to \( J_{\lambda} \), since \( \tau(\cdot) \) may not be convex, so only one side of the saddle point inequality holds. Substituting \( g = g_{0} \) into (\hyperref[e12]{12}), we obtain:
\[
J_{\lambda} \leq \mathcal{V}_{P,Q}(f_{\lambda}, g_{0}) + \lambda s(g_{0}).
\]
Taking the upper limit as \( \lambda \to 0 \), we have:
\[
\limsup_{\lambda \to 0} J_{\lambda} \leq \limsup_{\lambda \to 0} \left[ \mathcal{V}_{P,Q}(f_{\lambda}, g_{0}) + \lambda s(g_{0}) \right].
\]

By \hyperref[l4.1]{Lemma 4.1}, since $S(g)$ is bounded, \( \lim_{\lambda \to 0} \lambda S(g) = 0 \), so:
\[
\limsup_{\lambda \to 0} J_{\lambda} \leq \mathcal{V}_{P,Q}(f_{\lambda}, g_{0}). \tag{13} \label{e13}
\]
Substituting \( f = f_{\lambda} \) into (\hyperref[e11]{11}), we have:
\[
\mathcal{V}_{P,Q}(f_{\lambda}, g_{0}) \leq \mathcal{V}_{P,Q}(f_{0}, g_{0}) = J_{0}. \tag{14} \label{e14}
\]
Combining (\hyperref[e13]{13}) and (\hyperref[e14]{14}), we obtain:
\[
\limsup_{\lambda \to 0} J_{\lambda} \leq \mathcal{V}_{P,Q}(f_{\lambda}, g_{0}) \leq \mathcal{V}_{P,Q}(f_{0}, g_{0}) = J_{0}.
\]
Thus, with \( J_{\lambda} \geq J_{0} \) and \(\limsup_{\lambda \to 0} J_{\lambda} \leq J_{0} \), \hyperref[t4.3]{Theorem 4.3} is proved.
\hfill $\square$

\noindent \textbf{Proof of \hyperref[t4.4]{Theorem 4.4}:} After proving \hyperref[t4.3]{Theorem 4.3}, we aim to show that when \( \lambda \to 0 \),
\[
\
\mathrm{dist}\Bigl(\bigl(g_{\lambda}, f_{\lambda}\bigr),\, \mathcal{G}\Bigr) 
\;\rightarrow\; 0,
\]
Under \hyperref[a3.1]{Assumption 3.1}, the sets \(\mathcal{S}(P)\) and \(\mathcal{S}(Q)\) are uniformly compact and uniformly bounded in \(L^{\infty}\), with their essential supremum providing the uniform bound. Additionally, by \hyperref[l4.1]{Lemma 4.1}, all \(f \in \mathcal{S}(P)\) and \(g \in \mathcal{S}(Q)\) are \(L\)-\emph{Lipschitz} continuous, which implies that \(\mathcal{S}(P)\) and \(\mathcal{S}(Q)\) are \emph{equicontinuous}. 

The combination of uniform boundedness in \(L^{\infty}\) and \emph{equicontinuity} implies that \(\mathcal{S}(P)\) and \(\mathcal{S}(Q)\) satisfy the conditions of the Arzelà-Ascoli Theorem. Applying the theorem on arbitrary compact metric spaces, or more generally, on compact Hausdorff spaces, ensures the existence of a convergent subsequence.

Consider a sequence of optimal solutions \((f_{\lambda_n}, g_{\lambda_n}) \in \mathcal{S}(P) \times \mathcal{S}(Q)\) corresponding to a sequence \(\lambda_n \to 0\). Due to the compactness of \(\mathcal{S}(P) \times \mathcal{S}(Q)\), there exists a uniformly convergent subsequence \((f_{\lambda_{n_k}}, g_{\lambda_{n_k}}) \to (f', g')\), where \((f', g') \in \mathcal{S}(P) \times \mathcal{S}(Q)\). 

This result relies on the fact that the uniform boundedness in \(L^{\infty}\) ensures the subsequential limits retain their essential supremum bounds, and the equicontinuity guarantees convergence in the uniform topology. The completeness of \(L^{\infty}\) further ensures the limit functions \(f'\) and \(g'\) are well-defined within \(\mathcal{S}(P)\) and \(\mathcal{S}(Q)\), respectively, which validates the application of the Arzelà-Ascoli Theorem in this context \cite{dunford1958linear}.

Since \( \lim_{\lambda \to 0} J_{\lambda} = J_{0} \), we have:
\[
\lim_{n \to \infty} J_{\lambda_n} = J_{0}.
\]
Consequently,
\[
\lim_{n \to \infty} J_{\lambda_n} = \lim_{n \to \infty} \mathcal{V}_{P,Q}(f_{\lambda_n}, g_{\lambda_n}) + \lambda_{n} s(g_{\lambda_n}) = \mathcal{V}_{P,Q}(f', g').
\]
This implies that:
\[
\mathcal{V}_{P,Q}(f', g') = J_{0} = \mathcal{V}_{P,Q}(f_{0}, g_{0}).
\]
Since \( \mathcal{V}_{P,Q} (f,g) + \lambda s(g) \) is concave-convex when \( \tau(\cdot) \) is convex, saddle point inequality holds for both $J_{\lambda}$ and $J_{0}$. Note that such saddle point is unique given the objective function is strongly concave-strongly convex \cite{NEURIPS2020_331316d4}. Our objective function is only concave-convex, which indicates that the convergent subsequence would possibly converge to a set of saddle points $\mathcal{G}$.

Due to the continuity of $\mathcal{V}_{P,Q} (f,g)$, the set $\mathcal{G}$ is closed. Therefore, we conclude that:
\[
\lim_{\lambda \to 0} 
\mathrm{dist}\Bigl(\bigl(g_{\lambda}, f_{\lambda}\bigr),\, \mathcal{G}\Bigr) 
\;=\; 0,
\]
\noindent It is important to note that when \( \tau(\cdot) \) is non-convex, while a convergent subsequence may still exist, the convergence to \( (f_{0}, g_{0}) \) is not guaranteed to hold. \hfill $\square$

\textbf{Proof of \hyperref[t4.5]{Theorem 4.5}:} We first define:
\[
J_{\lambda_{1}} = C_{P,Q} + \sup_{f \in \mathtt{ICNN}(\mathbb{R}^d)} \inf_{g \in \mathtt{ICNN}(\mathbb{R}^d)} \mathcal{V}_{P,Q}(f, g) + \lambda_{1} s(g),
\]
\[
J_{\lambda_{2}} = C_{P,Q} +\sup_{f \in \mathtt{ICNN}(\mathbb{R}^d)} \inf_{g \in \mathtt{ICNN}(\mathbb{R}^d)}  \mathcal{V}_{P,Q}(f, g) + \lambda_{2} s(g).
\]
Consider $(f_{\lambda_{1}}, g_{\lambda_{1}})$ and  $(f_{\lambda_{2}}, g_{\lambda_{2}})$ be the optimal solution to respect $J_{\lambda_{1}}$ and $J_{\lambda_{2}}$, we aim to prove that given $\lambda_{1} < \lambda_{2}$, we have
\[s(g_{\lambda_{1}}) \geq s(g_{\lambda_{2}}).
\]
By the saddle point inequality, we have the following four inequalities, 
\[
\mathcal{V}_{P,Q}(f_{\lambda_{1}}, g_{\lambda_{1}}) + \lambda_{1} s(g_{\lambda_{1}}) \leq \mathcal{V}_{P,Q}(f_{\lambda_{1}}, g_{\lambda_{2}}) + \lambda_{1} s(g_{\lambda_{2}}),\label{e15} \tag{15}
\]
\[
\mathcal{V}_{P,Q}(f_{\lambda_{2}}, g_{\lambda_{2}}) + \lambda_{2} s(g_{\lambda_{2}}) \leq \mathcal{V}_{P,Q}(f_{\lambda_{2}}, g_{\lambda_{1}}) + \lambda_{2} s(g_{\lambda_{1}}), \label{e16}\tag{16}
\]
\[
\mathcal{V}_{P,Q}(f_{\lambda_{1}}, g_{\lambda_{1}}) + \lambda_{1} s(g_{\lambda_{1}}) \geq \mathcal{V}_{P,Q}(f_{\lambda_{2}}, g_{\lambda_{1}}) + \lambda_{1} s(g_{\lambda_{1}}), \label{e17} \tag{17}
\]
\[
\mathcal{V}_{P,Q}(f_{\lambda_{2}}, g_{\lambda_{2}}) + \lambda_{2} s(g_{\lambda_{2}}) \geq \mathcal{V}_{P,Q}(f_{\lambda_{1}}, g_{\lambda_{2}}) + \lambda_{2} s(g_{\lambda_{2}}). \label{e18}\tag{18}
\]

By (\hyperref[e17]{17}) and (\hyperref[e18]{18}), we have 
\[
\mathcal{V}_{P,Q}(f_{\lambda_{1}}, g_{\lambda_{1}}) \geq \mathcal{V}_{P,Q}(f_{\lambda_{2}}, g_{\lambda_{1}}), \label{e19} \tag{19}
\]
\[
\mathcal{V}_{P,Q}(f_{\lambda_{2}}, g_{\lambda_{2}}) \geq \mathcal{V}_{P,Q}(f_{\lambda_{1}}, g_{\lambda_{2}}). \label{e20}\tag{20}
\]

Rearranging (\hyperref[e15]{15}) and (\hyperref[e16]{16}), we have

\[
\mathcal{V}_{P,Q}(f_{\lambda_{1}}, g_{\lambda_{1}}) - \mathcal{V}_{P,Q}(f_{\lambda_{1}}, g_{\lambda_{2}}) +  \mathcal{V}_{P,Q}(f_{\lambda_{2}}, g_{\lambda_{2}}) - \mathcal{V}_{P,Q}(f_{\lambda_{2}}, g_{\lambda_{1}})\leq(\lambda_{1} -\lambda_{2})[s(g_{\lambda_{2}}) -  s(g_{\lambda_{1}})].
\]

With (\hyperref[e19]{19}) and (\hyperref[e20]{20}), we can get

\[
0\leq(\lambda_{1} -\lambda_{2})[s(g_{\lambda_{2}}) -  s(g_{\lambda_{1}})],
\]
\[
s(g_{\lambda_{1}}) \geq s(g_{\lambda_{2}}) \quad (\because \lambda_{1}<\lambda_{2}).
\]
\hfill $\square$

\textbf{Proof of \hyperref[t4.6]{Theorem 4.6}: }
We first define the optimality gaps $\widehat{\epsilon}_{1}$ and $\widehat{\epsilon}_{2}$ for unbiased formulation (\hyperref[e4]{4})
\[
\widehat{\epsilon}_1(f, g) = \mathcal{V}_{P,Q}(f, g) - \inf_{g' \in \mathcal{S}(Q)} \mathcal{V}_{P,Q}(f, g'),
\]
\[
\widehat{\epsilon}_2(f) = \mathcal{V}_{P,Q}(f_{0}, g_{0}) - \inf_{g' \in \mathcal{S}(Q)} \mathcal{V}_{P,Q}(f, g').
\]
Leveraging from Theorem 3.6 from \citet{Makkuva20}, we then aim to bound $\{\widehat{\epsilon}_{1}, \widehat{\epsilon}_{2}\}$ using  $\{\epsilon_{1}, \epsilon_{2}\}$ and therefore bounding the upper bound of the map error:
\begin{align*}
    \widehat{\epsilon}_1(f, g) &= \mathcal{V}_{P,Q}(f, g) - \inf_{g' \in \mathcal{S}(Q)} \mathcal{V}_{P,Q}(f, g') \\
    &\leq [\mathcal{V}_{P,Q}(f, g) + \lambda s(g)] - [\mathcal{V}_{P,Q}(f,f^*) + \lambda s(f^*)] + \lambda s(f^*)\\
    &\leq \mathcal{N}_{P,Q}(f, g) - \inf_{g' \in \mathcal{S}(Q)} \mathcal{N}_{P,Q}(f, g') + \lambda s(f^*) \\
    &\leq \epsilon_{1}(f,g) + \lambda M_{\tau} \\
\end{align*}Similarly,\begin{align*}
    \widehat{\epsilon}_2(f) &= \mathcal{V}_{P,Q}(f_{0}, g_{0}) - \inf_{g' \in \mathcal{S}(Q)} \mathcal{V}_{P,Q}(f, g') \\
    &\leq \mathcal{V}_{P,Q}(f_{\lambda}, g_{\lambda}) - \inf_{g' \in \mathcal{S}(Q)} \mathcal{V}_{P,Q}(f, g') \\
    &\leq [\mathcal{V}_{P,Q}(f_{\lambda}, g_{\lambda}) + \lambda s(g_{\lambda})] - [\lambda s(g') - \inf_{g' \in \mathcal{S}(Q)} \mathcal{V}_{P,Q}(f, g')] + \lambda s(g')\\
    &\leq [\mathcal{V}_{P,Q}(f_{\lambda}, g_{\lambda}) + \lambda s(g_{\lambda})] -  \inf_{\tilde{g} \in \mathcal{S}(Q)} \mathcal{N}_{P,Q}(f, \tilde{g}) + \lambda s(g')\\
    &\leq \epsilon_{2}(f)+\lambda M_{\tau}\\
\end{align*}
By Theorem 3.6 from \citet{Makkuva20}, we therefore have 
\[
\|\nabla g - \nabla g_{0}\|^2_{L^2(Q)} \leq \frac{2}{\alpha} \left( \widehat{\epsilon}_1 + \widehat{\epsilon}_2\right) \leq \frac{2}{\alpha} \left( \epsilon_1 + \epsilon_2 + 2\lambda M_{\tau} \right).
\]

\section{Heuristic Algorithm Design and Technical Details}\label{ac}
\textbf{Low-Dimensional Tasks.} We again walk through the two main function \textproc{SimulatedAnnealingProposal} (abbreviated as \textproc{SA} in following) and \textproc{ACCEPT} in \hyperref[a1]{Algorithm 1} with detail.

\textproc{SA} is designed to select a new $\lambda$ for the next iteration. In the initial stages, a higher temperature $T$ allows for a wider range of $\lambda$ values to be explored, which encourages exploration and helps prevent the algorithm from getting stuck in local optima early in training. As training progresses, the temperature gradually decreases, resulting in a narrower range for $\lambda$ to promote convergence.

For \textproc{Accept}, we accept $\lambda$ if it improves $\text{Eval}$, or—with probability depends on temperature $T$—even if it worsens it. This simulated annealing encourages exploration at high temperature $T$, and $T$ gradually decreases over training. In \textproc{Evaluation} function from \eqref{e8}, we normalize $\text{Spa}$ and $\text{Res}$ to the same numerical level according to their minimum and maximum attainable ranges respectively. This, along with the adaptive $\lambda$ objective function, helps avoid getting stuck in a suboptimal local minimum during the early stages of large-scale minimax training. 

\begin{figure}[h]
\centering
\begin{minipage}[t]{0.50\linewidth}
\begin{algorithm}[H]
\caption{\textproc{SimulatedAnnealingProposal}}
\label{a1}
\begin{algorithmic}[1]
\REQUIRE Searching radius $r$, original intensity $\lambda$, current temperature $\text{Tem}$, lower clipping $R_{\text{low}}$
\REQUIRE Dimensionality constraint $l$ (High-dimensional tasks only)
\STATE $R \gets \max\left\{R_{\text{low}}, \exp(-r\cdot(1 - \text{Tem}))\right\}$
\IF{Low-Dimensional Tasks}
    \STATE $\lambda_{\text{new}} \gets \lambda \cdot \left(1 + \mathrm{random}(-R, R)\right)$
\ELSIF{High-Dimensional Tasks}
    \IF{Not reaching $l$}
        \STATE $\lambda_{\text{new}} \gets \lambda \cdot \left(1 + \mathrm{random}(0, R)\right)$
    \ELSE
        \STATE $\lambda_{\text{new}} \gets \lambda \cdot \left(1 + \mathrm{random}(-R, 0)\right)$
    \ENDIF
\ENDIF
\RETURN $\lambda_{\text{new}}$
\end{algorithmic}
\end{algorithm}
\end{minipage}
\hfill
\begin{minipage}[t]{0.49\linewidth}
\begin{algorithm}[H]
\caption{\textproc{Accept}}
\label{alg:accept}
\begin{algorithmic}[1]
\REQUIRE Current evaluation $\text{Eval}_{\text{new}}$, previous evaluation $\text{Eval}$, current temperature \text{Tem}
\STATE $\delta_{\text{eval}} \gets \text{Eval}_{\text{new}} - \text{Eval}$
\IF{$\delta_{\text{eval}} > 0$ \textbf{or} $\mathrm{random}() < \exp\left(\delta_{\text{eval}} / \text{Tem}\right)$}
    \STATE \textbf{return} \texttt{accept}
\ELSE
    \STATE \textbf{return} \texttt{reject}
\ENDIF
\end{algorithmic}
\end{algorithm}
\end{minipage}
\end{figure}

\textbf{High-Dimensional Tasks.} In high-dimensional tasks, our method proceeds in two phases: we first increase $\lambda$ to enforce the dimensionality constraint by inducing sparsity in the displacement vectors. Once the constraint is satisfied, we then decrease $\lambda$ heuristically to reduce the regularization strength, allowing the model to learn a more feasible transport map. All these are achieved through the temperature-based heuristic: at high temperatures (early training), the adjustment range for $\lambda$ is wide, encouraging exploration across sparsity levels. As training progresses and temperature decreases, the adjustment radius narrows, enabling fine-tuned reductions in sparsity to converge toward the most feasible solution.

We present additional training trajectories of the adaptive-$\lambda$ scheme in \hyperref[ad1]{Appendix D.1}, evaluated on scRNA perturbation task. This adaptive control strategy enables the solver to more effectively approach a solution closer to the global optimum comparing the scheme with constant-$\lambda$ through the training. We include the hyperparameters for both low- and high-dimensional experiments in \hyperref[ae]{Appendix E}.

\begin{algorithm}[h]
\caption{High-Dimensional Simulated‑Annealing Heuristic for Adaptive Regularization $\lambda$}
\label{a4}
\begin{algorithmic}[1]
\REQUIRE Initialization iterations $n_{\text{ini}}$, training iterations $n_{\text{tr}}$, dimensionality constraint $l$
\REQUIRE Initial temperature $\text{Tem}$, minimum temperature $\text{Tem}_{\min}$, decay rate $d$, initial intensity $\lambda$

\STATE \textbf{Initialization:} Train $(f,g)$ with $\lambda$ for $n_{\text{ini}}$ iterations
\STATE $\text{Dim}_{T} \gets \textproc{ComputeDim}(T)$

\WHILE{$\text{Dim}_{T} \geq l$ \AND $\text{Tem}> \text{Tem}_{\min}$}
    \STATE $\lambda' \gets \textproc{SimulatedAnnealingProposal}(T, \lambda)$ \hfill \texttt{\# Increase $\lambda$ for sparsity}
    \STATE Train $(f, g)$ with $\lambda'$ for $n_{\text{tr}}$ iterations
    \STATE $\text{Dim}_{T} \gets \textproc{ComputeDim}(T)$
    \STATE $\text{Tem} \gets d \cdot \text{Tem}$
    \STATE $\lambda \gets \lambda'$ 
\ENDWHILE

\WHILE{$\text{Tem} > \text{Tem}_{\min}$}
    \STATE  $\lambda' \gets \textproc{SimulatedAnnealingProposal}(T, \lambda)$ \hfill \texttt{\# Decrease $\lambda$ for feasibility} 
    \STATE Train $(f, g)$ with $\lambda'$ for $n_{\text{tr}}$ iterations
    \STATE $\text{Dim}_{T} \gets \textproc{ComputeDim}(T)$
    \IF{$\text{Dim}_{T} \geq l$}
        \STATE $\lambda \gets \lambda'$ 
    \ENDIF
    \STATE $\text{Tem} \gets d \cdot \text{Tem}$
\ENDWHILE

\end{algorithmic}
\end{algorithm}

\section{Experiments}\label{ad}
\subsection{Adaptive-$\lambda$ Training Trajectories and Results}\label{ad1}

\begin{table*}[h]\label{t2}
\centering
\caption{Detailed performance comparison on the synthesized task. For adaptive method, we include the results for three consecutive independent runs.}
\label{tab:ot_comparison}
\resizebox{\textwidth}{!}{%
\begin{tabular}{lcc|ccccc|ccc}
\toprule
\textbf{Method} & \textbf{ICNN OT} & \textbf{Cuturi et al. 23} & \multicolumn{5}{c|}{\textbf{Ours (Constant $\lambda$)}} & \multicolumn{3}{c}{\textbf{Ours (Adaptive $\lambda$)}} \\
\cmidrule(lr){4-8} \cmidrule(lr){9-11}
& & & $5\!\times\!10^{-4}$ & $1\!\times\!10^{-3}$ & $5\!\times\!10^{-3}$ & $7\!\times\!10^{-3}$ & $1\!\times\!10^{-2}$ & Trial 1 & Trial 2 & Trial 3 \\
\midrule
\textbf{dim($\Delta(\mathbf{x})$)} & 2480 & 99 & 153 & 138 & 120 & 120 & 121 & 111 & 110 & 107 \\
\bottomrule
\end{tabular}%
}
\end{table*}

\begin{figure*}[htbp]
    \centering
    \begin{subfigure}{0.325\textwidth}
        \centering
        \includegraphics[width=\linewidth]{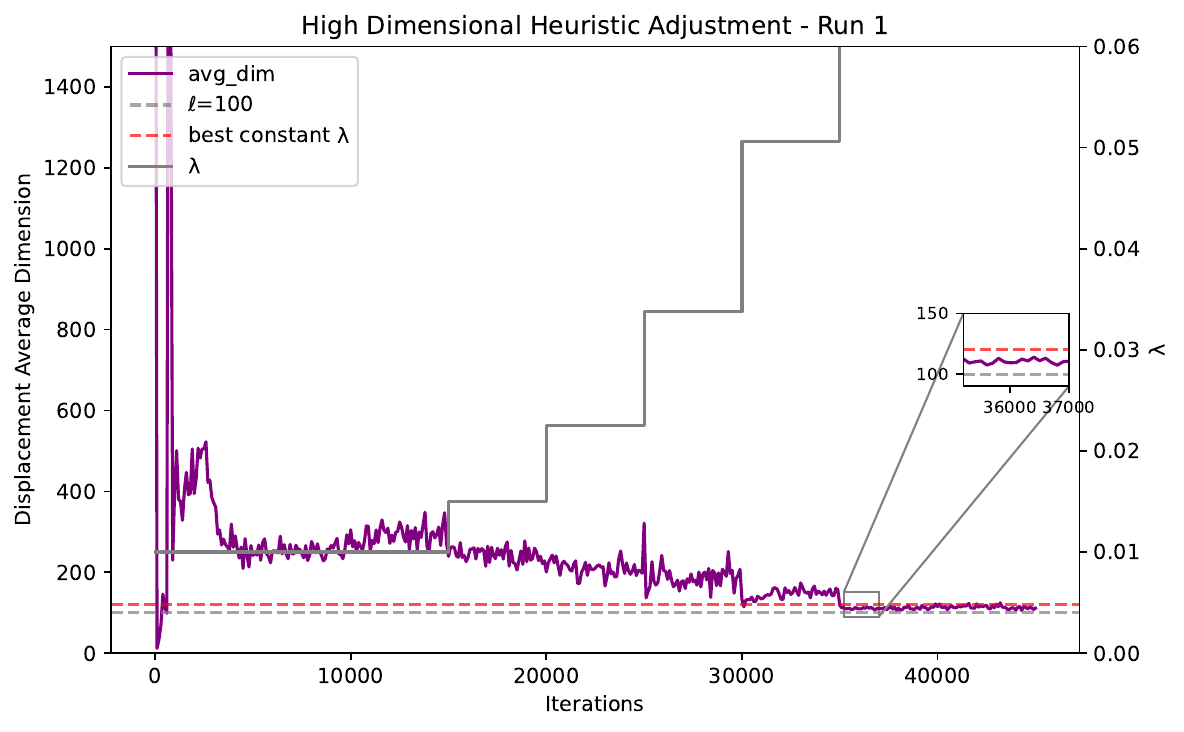}
        \caption{Trial 1: Final dim = 111}
    \end{subfigure}
    \hfill
    \begin{subfigure}{0.325\textwidth}
        \centering
        \includegraphics[width=\linewidth]{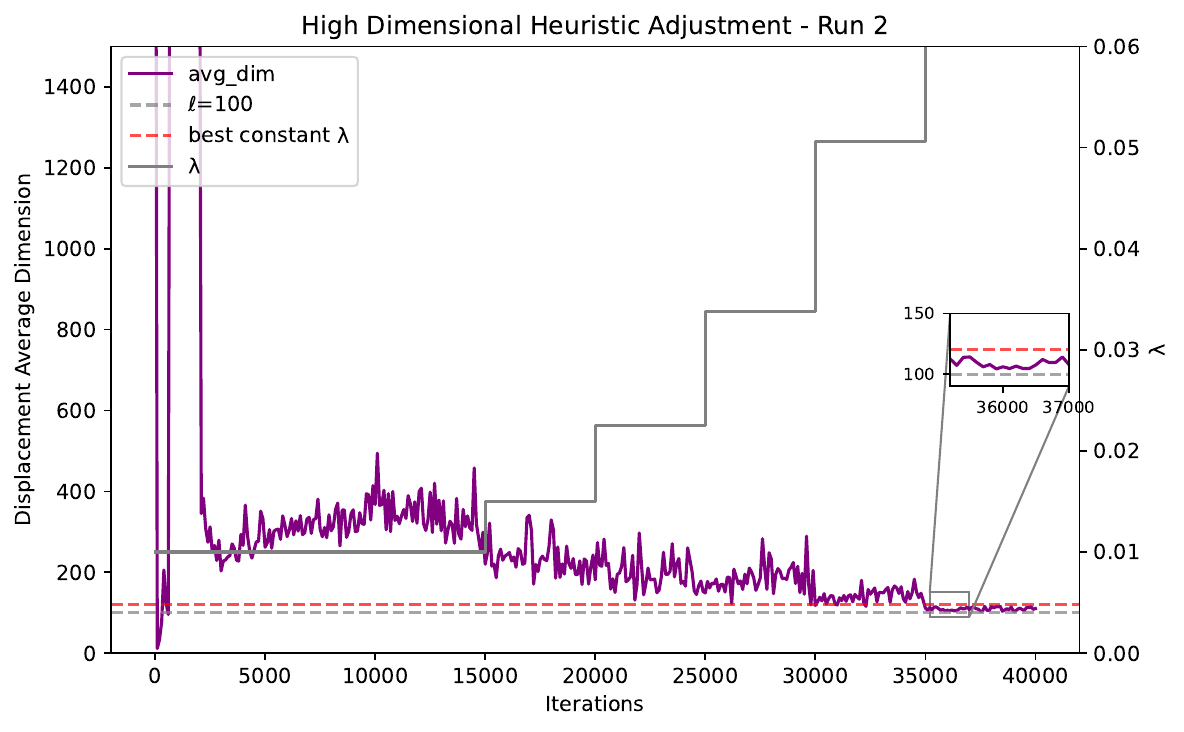}
        \caption{Trial 2: Final dim = 110}
    \end{subfigure}
    \hfill
    \begin{subfigure}{0.325\textwidth}
        \centering
        \includegraphics[width=\linewidth]{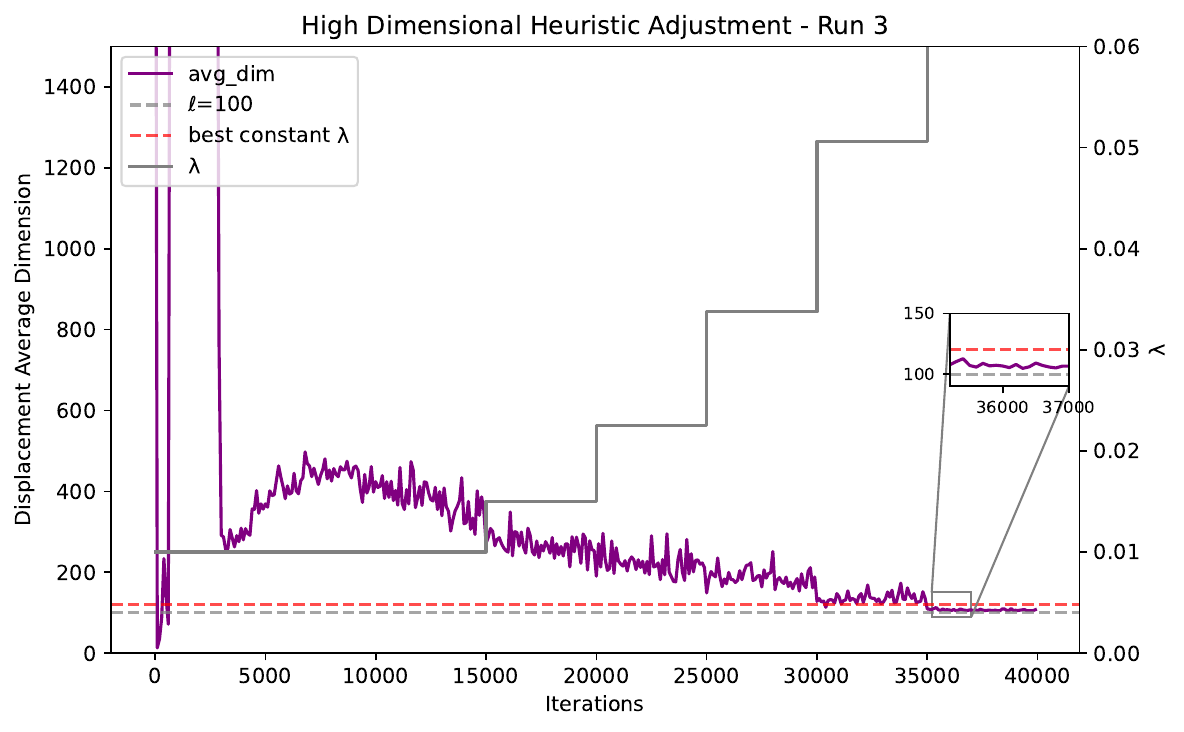}
        \caption{Trial 3: Final dim = 107}
    \end{subfigure}
    
    \caption{Three consecutive runs of high-dimensional dynamic adjustment of $\lambda$ under the same seed. Dimensionality constraint is set to $l = 100$ and smoothed $\ell_0$ penalty is used.}
    \label{f7}
\end{figure*}

As shown in \hyperref[t2]{Table 2}, we include additional results for both the constant and adaptive $\lambda$ schemes. We highlight the necessity of using the adaptive $\lambda$ scheme—as shown in \hyperref[f7]{Figure 7}—demonstrating that embedding the simulated annealing scheme into ICNN training yields robust results across multiple consecutive runs, consistently outperforming the constant scheme.

\subsection{Selected Genes Overlapping Accuracy}\label{ad2}

For experiments with known ground-truth dimensionality and selected genes, simply producing a dimensionality close to the ground truth is not sufficient. It is also necessary to verify whether the genes selected by the solver accurately overlap with the ground-truth perturbed genes. To evaluate robustness under varying levels of ground-truth dimensionality, we construct two additional datasets with $k = 100$ and $k = 500$, where $k$ denotes the number of truly perturbed genes.

\begin{table}[h]
\centering
\small
\caption{Effects of different $\lambda$ values on displacement dimensions and gene overlap accuracy across two synthesized datasets ($k = 100, 500$), where $k$ denotes the number of ground-truth low-dimensional genes. Each value is averaged over three consecutive runs.}
\vspace{1em}
\begin{tabular}{@{}c c c c@{}}
\toprule
\textbf{$k$} & \textbf{$\lambda$} & \textbf{$\text{dim}(\Delta(\mathbf{x}))$} & \textbf{Overlap \% with Ground Truth} \\
\midrule
\multirow{4}{*}{500} 
 & 0.0005   & $532.7 \pm 8.6$   & $0.9973 \pm 0.0023$ \\
 & 0.001    & $519.7 \pm 5.5$   & $0.9953 \pm 0.0012$ \\
 & 0.005    & $505.0 \pm 5.6$   & $0.9893 \pm 0.0076$ \\
 & Adaptive & $501.3 \pm 2.7$   & $0.9913 \pm 0.0012$ \\
\midrule
\multirow{4}{*}{100} 
 & 0.0005   & $160.3 \pm 7.4$   & $1.0000 \pm 0.0000$ \\
 & 0.001    & $142.3 \pm 3.0$   & $0.9633 \pm 0.0208$ \\
 & 0.005    & $120.7 \pm 4.1$   & $0.9500 \pm 0.0100$ \\
 & Adaptive & $109.3 \pm 2.1$   & $0.9567 \pm 0.0251$ \\
\bottomrule
\end{tabular}
\label{t3}
\end{table}

\begin{figure*}[!htbp]
    \centering
    \includegraphics[width=\textwidth]{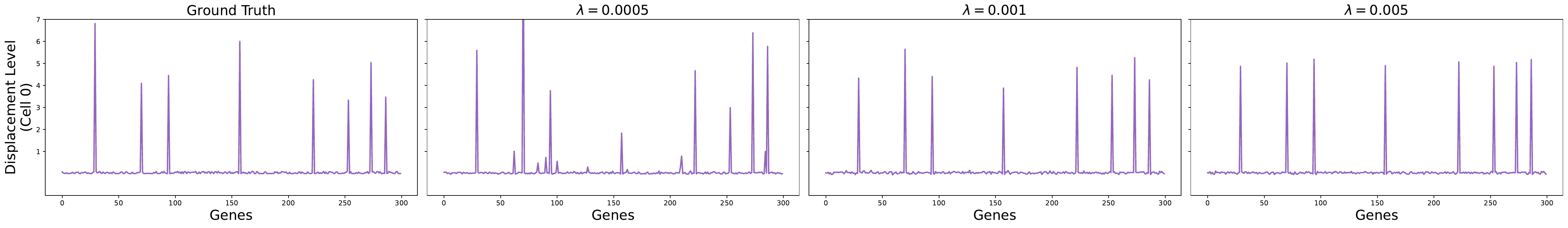} 
    \vspace{-1em} 
    \caption{Examples of displacement vectors for the first 300 genes on the first cells in the $k = 100$ dataset. The first column displays the ground truth displacement vectors from \textit{Cell}~0; the following columns show the displacement vectors from the \textit{Cell}~0 for different values of~$\lambda$.}
    \label{f8}
\end{figure*}

\begin{figure*}[!htbp]
    \centering
    \includegraphics[width=\textwidth]{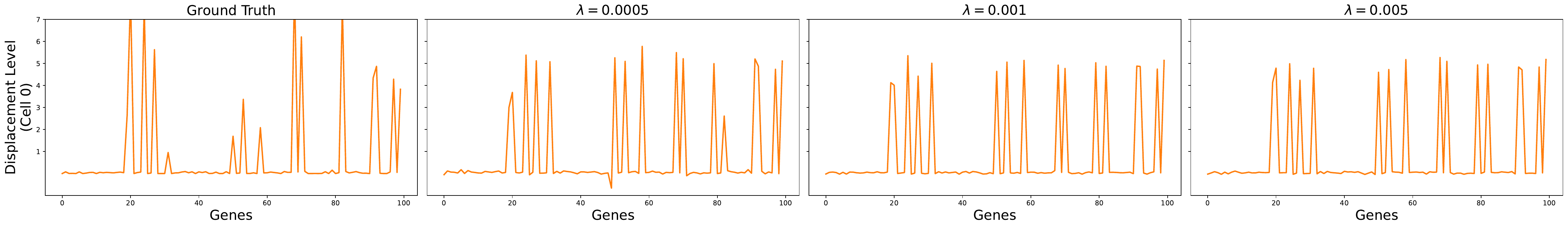} 
    \vspace{-1em} 
    \caption{Examples of displacement vectors for first 100 genes on \textit{Cell}~0 in the $k=500$ dataset.}
    \label{f9}
\end{figure*}

As shown in \hyperref[t3]{Table 3}, our method achieves competitive accuracy in recovering the truly perturbed genes. To further illustrate how genes are displaced during perturbation, we include examples of gene displacement vector under different sparsity intensities $\lambda$. As shown in \hyperref[f8]{Figure 8} and \hyperref[f9]{Figure 9}, we visualize the gene displacement for the first cell in the experiment. At lower sparsity levels, the resulting map tends to be noisy comparing to the ground-truth map, including non-perturbed but noisy genes. As sparsity increases, the map becomes cleaner and more focused on the truly perturbed genes.

\subsection{Real 4i Experiment}\label{ad3}

This task assesses the effectiveness of various methods in recovering the OT map, focusing on preserving key features while disregarding less significant ones. Unlike sc-RNA perturbation, where there is a clear separation between noisy and genuinely perturbed cells, recovering the OT map is notably more challenging for 4i case. Additionally, the absence of a ground-truth map with the true dimensionality further complicates the task.

\textbf{Experiment Setup Details.} Datasets used in this experiment are publicly available at the following \href{https://www.research-collection.ethz.ch/handle/20.500.11850/609681}{site} \cite{20.500.11850/609681}. The setup follows from the sc-RNA experiment: after drug perturbation, the 4i features of the cells are altered, and we use these features to match control cells with perturbed cells. We selected four drugs (Ixazomib, Sorafenib, Cisplatin, Melphalan) to highlight the improvement of our method towards the existing methods.

\hyperref[f6]{Figure 6} shows the results of the 4i perturbation under these four drugs. Each experiment, conducted at a specific $\lambda$, displays the average outcome of $10$ consecutive independent runs. $\lambda^{(i)}$ represents the ordered statistics of each set of $\lambda$ used in the experiments. Since perturbation intensity varies by drug, which makes different $\lambda$ across different drugs necessary. \texttt{np.isclose(.,0)} is evaluated at the threshold of $10^{-2}$ to determine the dimensionality of the displacement vectors. Number of cells, $n$, in each experiment is approximately $12,000$ cells. We include all hyperparameters in \hyperref[ae]{Appendix E}. As shown in \hyperref[f6]{Figure 6}, among all four drugs, our method could more effectively reduce the dimensionality of the displacement vector to a even sparse level.

\begin{figure*}[!htbp]
    \centering
    \includegraphics[width=0.63\textwidth]{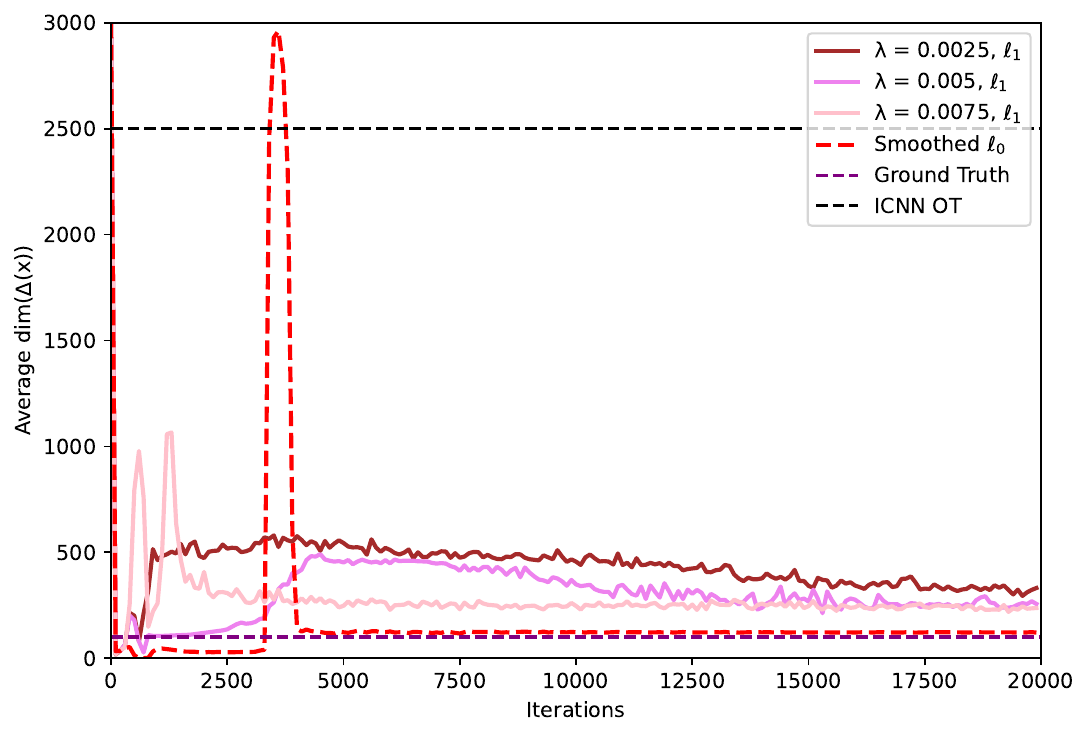} 
    \vspace{-0.9em}
    \caption{Different trajectories of $\lambda$ for the $\ell_1$ penalty are shown. The experiment is conducted on the dataset with a ground truth dimension of $k = 100$. ``Smoothed~$\ell_0$'' represents the training trajectory of the previously used Smoothed~$\ell_0$ with $\lambda = 0.005$.
}
    \label{f10}
\end{figure*}

\textbf{Further Comparison between $\ell_{0}$ and $\ell_{1}$.} We apply both $\ell_{0}$ and $\ell_{1}$ on the synthetic sc-RNA dataset with $k=100$. Shown in \hyperref[f10]{Figure 10}, the best achievable dimensionality of $\ell_{1}$-norm in neural OT setting is around 250, which is significantly higher than the one produced by the $\ell_{0}$-norm ($120$). This result highlights the effectiveness of $\ell_{0}$-norm to achieve more interpretable map in neural OT solvers.

\subsection{Limitations}

\textbf{Variance of the Results.} Due to the inherent randomness in ICNN, the results typically exhibit higher instability as $\lambda$ increases, leading to more varied outcomes across runs. This aligns with \hyperref[t4.6]{Theorem 4.6}, which larger $\lambda$ would result in a higher error bound. However, this error bound is still controllable through adaptive adjusting the intensity and more iterations of training that pushes optimality gap to $0$.

\textbf{Unknown and Varied Minimized Dimensionality.} The minimized dimensionality in each experiment varies and is data-driven, making it difficult and impractical to estimate a theoretical or empirical minimum. This highlights the necessity for our adaptive-$\lambda$ framework: if the target dimensionality in unattainable, then solver will adaptively adjust intensity and provide the best dimensionality it can achieve.

\textbf{Non-convex Penalty under Theoretical Guarantee.} Our theoretical guarantee (\hyperref[t4.4]{Theorem 4.4}, \hyperref[t4.5]{Theorem 4.5}) only applies to convex penalties. However, the smoothed $\ell_{0}$-norm is pseudoconvex and therefore does not benefit from the theoretical guarantee. This can be observed in the dimensionality for the highest $\lambda$ in each experiment, where the average dimensionality begins to increase. Non-convexity might be a contributing factor, but it is worth noting that the metric used to measure sparsity in \hyperref[t4.5]{Theorem 4.5} is the penalty function $\tau$, whereas we use the direct dimensionality metric in the experiment.

\section{Hyperparameters}\label{ae}

All the training of neural OT solvers in this work is conducted on \texttt{Intel8480-112C-2048G} \textbf{CPU only}. The training time for high-dimensional synthetic dataset is around 3 hours for constant-$\lambda$ and 4 hours for the adaptive-$\lambda$. This highlights the efficiency and applicability of our work for end-users from biological field without getting access to HPC GPU resources.

\textbf{Hyperparameters for \hyperref[f4]{Figure 4}.} We use most of the default hyperparameters for ICNN training. For iteration-related hyperparameters, we set $n_{\text{ini}}=20,000$ and $n_{\text{tr}}=n_{\text{sm}}=2,000$. For simulated annealing hyperparameters, we set the initial temperature to 1.0, the minimum temperature to 0.15, the temperature decay rate to 0.95, and the range adjustment parameter $p = 3$. For the $\tau_{\text{stvs}}$ penalty, we set $\gamma=100$.

\textbf{Hyperparameters for \hyperref[f5]{Figure 5}.} For the ICNN OT results, we largely use the default hyperparameter settings outlined in the original paper, with a modification to the batch size, which is set to 128 to better accommodate our datasets. For our results, we also use the same set of default parameters as ICNN OT. The parameter for the smoothed $\ell_{0}$-norm is set to $\sigma = 1.0$. For the results from \citet{Cuturi23}, we set the entropic regularization to $\epsilon=10^{-3}$ and the sparsity regularization intensity to $\lambda=2$.

\textbf{Hyperparameters for \hyperref[f6]{Figure 6}.} The default training parameters for ICNN are used here as well. We primarily experiment with different sparsity regularization intensities $\lambda$. For all four drugs, we use the same set of parameters $\lambda^{(i)}=[0.01, 0.1, 0.2, 0.5, 0.8, 1, 2]$ for the results from \citet{Cuturi23}, as the outcomes do not vary significantly across experiments. The dimensionality, however, starts to increase as $\lambda$ approaches 1. For our results, we find that each experiment varies under different parameter settings. Therefore, we adjust the parameters for each experiment to better illustrate how the dimensionality changes with different intensities.

\begin{itemize}
    \item For Ixazomib, we use $\lambda^{(i)}=[0.0001, 0.0005, 0.001, 0.0015, 0.002, 0.0025, 0.005]$.
    \item For Sorafenib, we use $\lambda^{(i)}=[0.001, 0.0015, 0.002, 0.0025, 0.003, 0.004, 0.005]$.
    \item For Cisplatin, we use $\lambda^{(i)}=[0.0001, 0.0005, 0.001, 0.0015, 0.0025, 0.004, 0.005]$.
    \item For Melphalan, we use $\lambda^{(i)}=[0.0002, 0.0005, 0.002, 0.0025, 0.0035, 0.004, 0.005]$.
\end{itemize}

\end{document}